\documentclass[runningheads]{llncs}

\usepackage{eccv}

\usepackage{eccvabbrv}

\usepackage{graphicx}
\usepackage{booktabs}

\usepackage{dsfont}

\usepackage{booktabs}
\usepackage{multirow} 
\usepackage{tabularx}
\usepackage{bm}
\usepackage{siunitx}
\usepackage{soul}
\usepackage{enumitem}

\usepackage{algorithm}
\usepackage{algorithmic}

\usepackage{marvosym}
\usepackage{wrapfig}

\usepackage[accsupp]{axessibility}  

\usepackage[pagebackref,breaklinks,colorlinks,citecolor=eccvblue]{hyperref}

\usepackage{orcidlink}

\begin{document}

\title{Distribution Alignment for Fully Test-Time Adaptation with Dynamic Online Data Streams} 

\titlerunning{DA-TTA}

\author{Ziqiang Wang\inst{1}\orcidlink{0000-0002-4083-5411} \and
Zhixiang Chi\inst{2}\orcidlink{0000-0003-4560-4986} \and
Yanan Wu\inst{3}\orcidlink{0000-0002-3301-6303} \and
Li Gu\inst{1}\orcidlink{0000-0002-4447-4967} \and
Zhi Liu\inst{4}\textsuperscript{\Letter}\orcidlink{0000-0002-8428-1131} \and
Konstantinos Plataniotis\inst{2}\textsuperscript{\Letter}\orcidlink{0000-0003-3647-5473} \and
Yang Wang\inst{1}\textsuperscript{\Letter}\orcidlink{0000-0001-9447-1791}}

\authorrunning{Z.~Wang et al.}

\institute{Concordia University, Canada \\
\email{\{ziqiang.wang,li.gu\}@mail.concordia.ca, yang.wang@concordia.ca} \and
University of Toronto, Canada \\
\email{zhixiang.chi@mail.utoronto.ca, kostas@ece.utoronto.ca} \and
Beijing Jiaotong University, China
\email{ynwu0510@bjtu.edu.cn} \and
Shanghai University, China 
\email{liuzhisjtu@163.com}
}

\maketitle
\let\oldthefootnote\thefootnote
\def\thefootnote{\Letter}\footnotetext{Corresponding authors. Our code is available at \href{https://github.com/WZq975/DA-TTA}{github.com/WZq975/DA-TTA}.}
\let\thefootnote\oldthefootnote
\setcounter{footnote}{0}

\begin{abstract}
  Given a model trained on source data, Test-Time Adaptation (TTA) enables adaptation and inference in test data streams with domain shifts from the source. Current methods predominantly optimize the model for each incoming test data batch using self-training loss. While these methods yield commendable results in ideal test data streams, where batches are independently and identically sampled from the target distribution, they falter under more practical test data streams that are not independent and identically distributed (non-i.i.d.). The data batches in a non-i.i.d. stream display prominent label shifts relative to each other. It leads to conflicting optimization objectives among batches during the TTA process. Given the inherent risks of adapting the source model to unpredictable test-time distributions, we reverse the adaptation process and propose a novel Distribution Alignment loss for TTA. This loss guides the distributions of test-time features back towards the source distributions, which ensures compatibility with the well-trained source model and eliminates the pitfalls associated with conflicting optimization objectives. Moreover, we devise a domain shift detection mechanism to extend the success of our proposed TTA method in the continual domain shift scenarios. Our extensive experiments validate the logic and efficacy of our method. On six benchmark datasets, we surpass existing methods in non-i.i.d. scenarios and maintain competitive performance under the ideal i.i.d. assumption.
\end{abstract}

\section{Introduction}

\begin{figure}[h]
    \centering
    \begin{subfigure}{0.42\textwidth}
        \includegraphics[width=\linewidth]{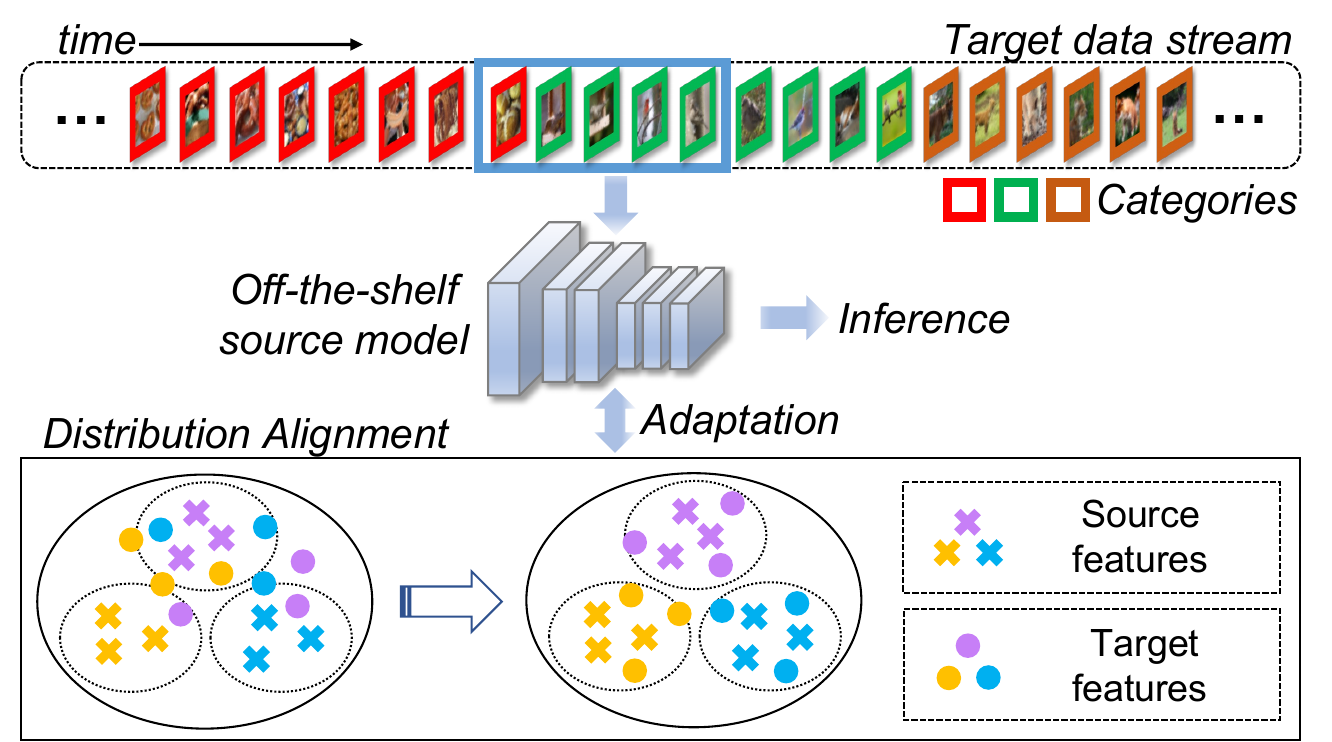}
        \caption{}
        \label{fig:intro_overview}
    \end{subfigure}
    \hspace{0.5cm}
    \begin{subfigure}{0.47\textwidth}
        \includegraphics[width=\linewidth]{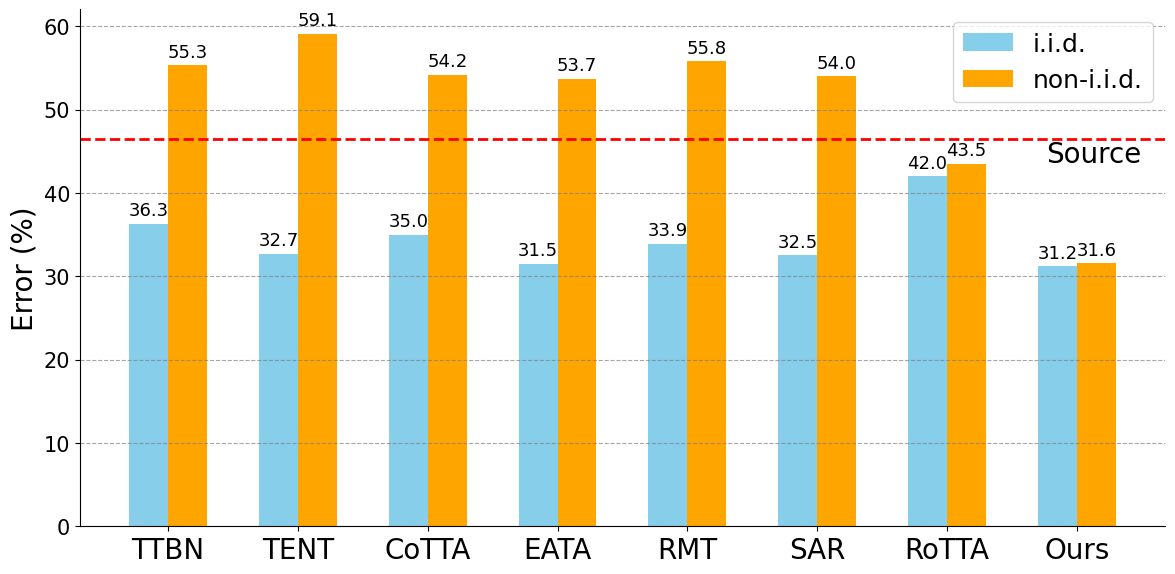}
        \caption{}
        \label{fig:intro}
    \end{subfigure}
    \caption{(a) Overview of our method for fully TTA. With each time step, there are distribution shifts (label shifts). Our proposed method aligns the distributions of test-time features with those of the source, not only mitigating the domain shift, but ensuring robust TTA in the non-i.i.d. data streams. (b) Average classification errors of TTA methods from CIFAR100 (source) to CIFAR100-C (target) under the fully TTA with i.i.d and non-i.i.d data stream settings. Lower is better. The red ``Source'' line indicates the source model's result without adaptation. Label shifts in the non-i.i.d. test data stream degrade TTBN and self-training based methods (SAR and RoTTA are designed for non-i.i.d. data streams).}
\end{figure}

The unprecedented success of deep models~\cite{he2016deep,ye2024bayes,yang2023conditional,chi2020all} is conditioned on the assumption that the training and test data are drawn from the same distribution~\cite{wang2021tent}. However, such an assumption is delicate in ever-changing deployment environments~\cite{koh2021wilds,fernando2013unsupervised}, leading to domain shift and performance deterioration.

Test-Time Adaptation (TTA) is a line of research that mitigates domain shift by continually adapting to the unlabeled data stream in a target domain before inference~\cite{wang2021tent,sun2020TTT,chi2021test,liang2020we,chiadapting}. There are two main categories of TTA: 1) Test-Time Training (TTT)~\cite{sun2020TTT}, where customized model training (\textit{e.g.}, adding auxiliary tasks~\cite{liu2023meta,wu2023metagcd})  is performed offline using the source data and performs adaptation on test data. 2) Fully TTA~\cite{wang2021tent}, adapts an off-the-shelf model without altering offline training. Our work focuses on fully TTA which poses a greater challenge as minimal training is allowed. 

There have been a series of studies on fully TTA  to tackle the challenges of learning on unlabeled data. One notable method is Test-time Batch Normalization (TTBN)~\cite{nado2020TTBN, schneider2020TTBN,wu2024test}. TTBN adjusts the BN layers, allowing them to normalize the feature leveraging the \textit{batch statistics from the current test data batch}, rather than the population statistics from the training phase. Given its effectiveness, TTBN has become a cornerstone for recent TTA research. Following this, there has been a surge in self-training-based TTA methods, primarily hinging on two adaptation objectives. The first employs entropy minimization (EM)~\cite{wang2021tent, niu2022EATA, 2022MEMO, gong2022NOTE, song2023ecotta, zhao2023delta}, pushing the model to make predictions with low entropy for the incoming test data. This ensures that the inference is well-distanced from the classification boundary~\cite{shu2018dirt}, thereby enhancing model performance on the test data. The second utilizes a teacher-student self-training framework~\cite{wang2022CoTTA, dobler2023RMT, yuan2023RoTTA,zhong2022meta}. Here, a teacher model assigns pseudo labels to the student, allowing the latter to be trained in a manner similar to supervised learning.

Both TTBN and self-training-based TTA methods aim to tailor the model \textit{towards the incoming unknown test data batch}. Their notable performances have been observed in ideal circumstances where every test data batch is independent and identically distributed (i.i.d., balanced classes) from the target domain. In the online fashion, as the model keeps updating, the i.i.d scenario ensures stable optimization from batch to batch. However, real-world scenarios are rarely so accommodating. In applications like self-driving vehicles and robot-vision systems, the batches of image feed are temporally correlated, making it non-i.i.d (imbalanced classes)~\cite{gong2022NOTE,Shayan2024Fed}. In the non-i.i.d. data streams, the long-tailed problem may occur, where a minority of classes dominate the current batch~\cite{zhang2023long_tailed} (see \cref{fig:intro_overview}). As each incoming batch of data contains different sets of dominated classes, the distributions among batches are diverse, leading to conflicting optimization. As depicted in \cref{fig:intro}, for TTBN, the adaptation fails because test data batches in long-tailed distributions do not provide true target domain statistics for the BN layers. For self-training-based methods, the varying long-tailed target distributions in TTA sessions lead to conflicting optimization objectives~\cite{kundu2022balancing}, which severely impair model performance, and may even cause it to collapse~\cite{Yang2024Markov,Yang2024how}.

To address the aforementioned challenges, we propose a surprisingly simple yet effective method for robust TTA, as illustrated in \cref{fig:intro_overview}. Rather than adapting the model to unpredictable test-time distributions, we reverse such a process and propose to set the source feature distribution as a reference and pull test data towards it. As a result, conflicting optimization objectives among data batches can be alleviated. Specifically, we propose to optimize the source model's affine layers using a Distribution Alignment (DA) loss. This loss minimizes the divergence between test feature distributions and the source distributions, thereby ensuring the test data's features align with the source model for compatibility. Furthermore, to accommodate scenarios featuring continuous domain shifts in test data streams, namely continual TTA~\cite{wang2022CoTTA, dobler2023RMT, yuan2023RoTTA}, we devise a domain shift detection mechanism that tracks changes in feature distributions. It improves our TTA method's efficacy in continuous domain environments. As demonstrated in \cref{fig:intro}, our method outshines others in handling both i.i.d. and non-i.i.d. data streams, effectively navigating the challenges associated with non-i.i.d. streams.

\textbf{Main Contributions:} (1) Our distribution alignment loss addresses the TTA challenges in non-i.i.d. scenarios by aligning test features with source distributions, ensuring they mesh with the source model and preventing degradation from conflicting optimization objectives. (2) We propose a domain shift detection mechanism that tracks feature distributions, enhancing our TTA method's performance for continual TTA in non-i.i.d. data streams. (3) Our method surpasses recent state-of-the-art methods (\eg, $\sim6\%$ on ImageNet-C/CIFAR100-C) across six datasets with different types of domain shifts in non-i.i.d. scenarios, while maintaining comparable performance under i.i.d. assumption.

\section{Related Work}
\noindent \textbf{Unsupervised Domain Adaptation (UDA).} 
UDA addresses the distribution shift by jointly training on the labeled source and unlabeled target data~\cite{wilson2020survey,li2021transferable,chen2018domain,liang2020we,kurmi2021domain}. One popular approach is to learn domain-invariant features by minimizing a certain measure of divergence between the source and target distributions (\textit{e.g.}~\cite{lin2022prototype,kang2019contrastive, long2015learning,sun2016deep,peng2019moment}). Another line of studies involves embedding a ``domain discriminator" within the network, which is applied to develop indistinguishable feature space (\textit{e.g.}~\cite{long2018conditional,ganin2015unsupervised,purushotham2016variational,pei2018multi}). However, the necessity of having access to both source and target domains during training limits the usability of these methods.

\noindent \textbf{Source-free Domain Adaptation (SFDA).} SFDA aims to adapt source models to unlabeled target domains without accessing the source domain data~\cite{wang2022exploring,huang2021model,yang2021generalized,ahmed2023ssda,tang2023consistency}. Among these, SHOT~\cite{liang2020we} suggests learning target-specific features through information maximization and pseudo-label prediction. 
SFDA-DE~\cite{ding2022source} works on domain alignment by estimating source class-conditioned feature distribution and minimizing a contrastive adaptation loss. DSiT~\cite{sanyal2023domain} utilizes the queries of a vision transformer to induce
domain-specificity and train the unified model to enable a disentanglement of task- and domain-specificity.
Most existing source-free methods~\cite{wang2022exploring, yang2022attracting, kundu2022balancing} operate offline and require an analysis of the entire test dataset, along with several adaptation epochs for model updates. Specially, BUFR~\cite{eastwood2022bufr} pre-computes and stores marginal distributions for each feature on source data using a soft binning function. It then realizes adaptation by restoring the test features with the stored marginal distributions. The philosophy of BUFR can be related to our work, thus, we also include a comparison with this SFDA method in our experiments.

\noindent \textbf{Test-time Adaptation (TTA).}
TTA can be categorized into Test-Time Training~\cite{sun2020TTT} (TTT) and Fully TTA~\cite{wang2021tent}, differentiated by the presence of prior joint training. TTT leverages both supervised and self-supervised losses to train a source model, which is then fine-tuned during TTA using self-supervised learning~\cite{liu2021ttt++, bartler2022mt3}. Fully TTA, in contrast, performs inference and adaptation directly in test data streams without prior training. A notable method in this setting is Test-Time Batch Normalization (TTBN)~\cite{nado2020TTBN, schneider2020TTBN}, which utilizes test-time batch statistics within BN layers for adaptation. Subsequently, optimization-based methods, including entropy minimization~\cite{wang2021tent, niu2022EATA, 2022MEMO, gong2022NOTE, song2023ecotta, zhao2023delta} and teacher-student self-training\cite{wang2022CoTTA, dobler2023RMT, yuan2023RoTTA, marsden2022gtta, marsden2023boid}, have been developed.
EATA~\cite{niu2022EATA} alleviates redundant optimization in test streams by employing a mechanism that identifies redundant samples. It tracks model outputs and skips the optimization for samples that are similar to previous ones. Our domain shift detection mechanism also adopts a tracking philosophy, albeit with a different focus and objective: to monitor feature distribution and detect domain shifts.
Besides, LAME~\cite{boudiaf2022LAME} focuses on adjusting output assignments rather than tuning parameters for TTA, while ODS~\cite{zhou2023ods} optimizes the estimation of label distribution to enhance self-training-based TTA methods in scenarios involving label shift. Moreover, DDA~\cite{gao2023dda} employs diffusion models to align target images with the source domain, then realizes classification without adapting the source model. 

\noindent \textbf{Modified TTBN for TTA in Non-i.i.d. Streams.}
TTBN~\cite{nado2020TTBN, schneider2020TTBN} establishes a strong baseline for TTA, yet it encounters difficulties in non-i.i.d. data streams or when dealing with small batch sizes. This is because incoming batches are class-imbalanced and provide biased statistics for BN layers. Subsequent works have modified TTBN to better handle non-i.i.d. streams or limited batch sizes. MEMO~\cite{2022MEMO} and TTN~\cite{lim2023ttn} combine source population statistics with dynamic test batch statistics, while DELTA~\cite{zhao2023delta} applies a moving average of test batch statistics for batch normalization. Furthermore, NOTE~\cite{gong2022NOTE} adjusts the normalization layers of TTBN by selectively incorporating instance normalization. In addition, both NOTE~\cite{gong2022NOTE} and RoTTA~\cite{yuan2023RoTTA} employ a resampling memory bank that collects and stores test samples from different estimated classes and updates test batch statistics from the stored samples using a moving average. 

Overall, prior works modify the computing of batch normalization, trying to stabilize the normalization process for the incoming test batch. Differently, this work investigates the correlation between model accuracy and the change in intermediate feature distribution due to imbalanced or balanced classes (\cref{sec:3.2}). And we introduce our Distribution Alignment method, which directly optimizes the distribution of features for all test batches towards the same source reference.

\section{Methodology}
\subsection{Problem Definition}
\label{sec:problem}

\begin{figure}[t]
    \centering
    \includegraphics[width=0.6\linewidth]{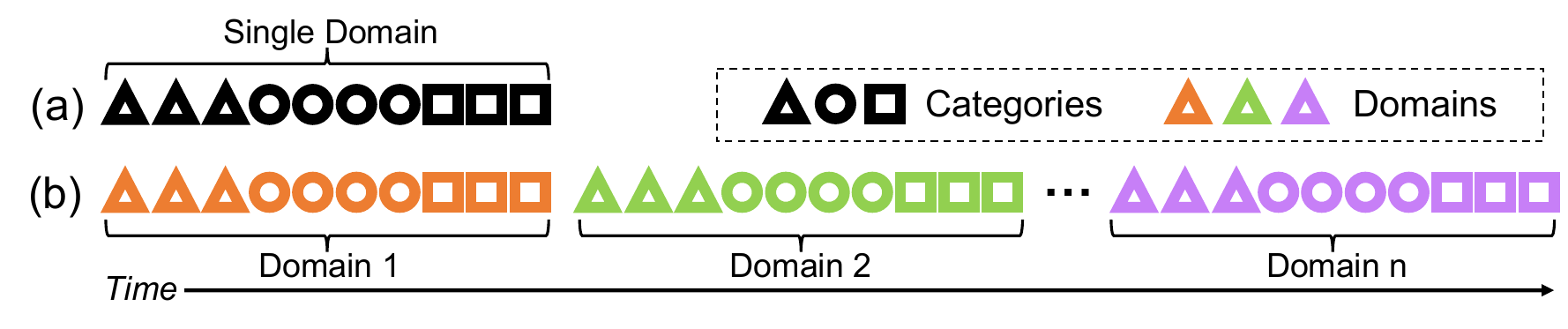}
    \caption{(a) Non-i.i.d. data streams. (b) Continual TTA with non-i.i.d. data streams.}
    \label{fig:stream}
\end{figure}

\textbf{Fully TTA}~\cite{wang2021tent} encompasses a scenario where a model, pre-trained on a labeled source dataset $\{(x, y) \sim P_S(x, y)\}$, is subjected to a stream of unlabeled test data from a target dataset $\{(x, y) \sim P_T(x, y)\}$. This target dataset presents a domain shift from the source, indicated by $P_S(x) \neq P_T(x)$ and $P_S(y|x) = P_T(y|x)$~\cite{storkey2009domainshift}. After deployment, the model updates itself based on the current data it receives, without using the source data. In pioneer work, fully TTA assumes that the distributions of target data over time, $P_T(x, y\mid t)$, are i.i.d. that is consistent with $P_S(x, y)$. However, our focus is on practical scenarios where $P_T(x, y\mid t)$ is \textbf{non-i.i.d.} and changes over time.
Therefore, \textbf{fully TTA in non-i.i.d. data streams} demands the management of both the domain shift from source to target and the distribution shifts (label shifts) that occur at each time step.
Besides, we also consider \textbf{continual TTA}~\cite{song2023ecotta, dobler2023RMT, yuan2023RoTTA}. This setting extends the fully TTA from a single target domain to a sequence of continuously shifting target domains:
\( P_{T_{1}}(x),\allowbreak P_{T_{2}}(x), \dots, \allowbreak P_{T_{n}}(x)\allowbreak \), as depicted in \cref{fig:stream}.

\subsection{Motivation on TTA in Dynamic Online Settings}
\label{sec:3.2}

TTBN~\cite{nado2020TTBN, schneider2020TTBN} sets a strong baseline underpinning a series of TTA works~\cite{wang2021tent, niu2022EATA, dobler2023RMT, wang2022CoTTA, marsden2022gtta, marsden2023boid, lim2023ttn}. We first analyze why TTBN experiences performance drop in dynamic online data streams follow by our motivation.

\begin{figure}[t]
    \centering
    \begin{subfigure}{0.2\textwidth}
        \includegraphics[width=\linewidth]{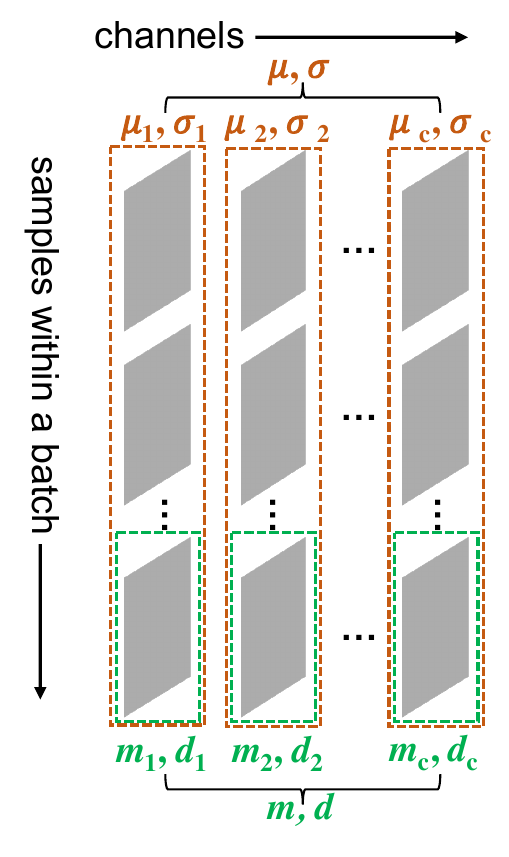}
        \caption{}
        \label{fig:muVSm}
    \end{subfigure}
    \hspace{0.5cm}
    \begin{subfigure}{0.5\textwidth}
        \includegraphics[width=\linewidth]{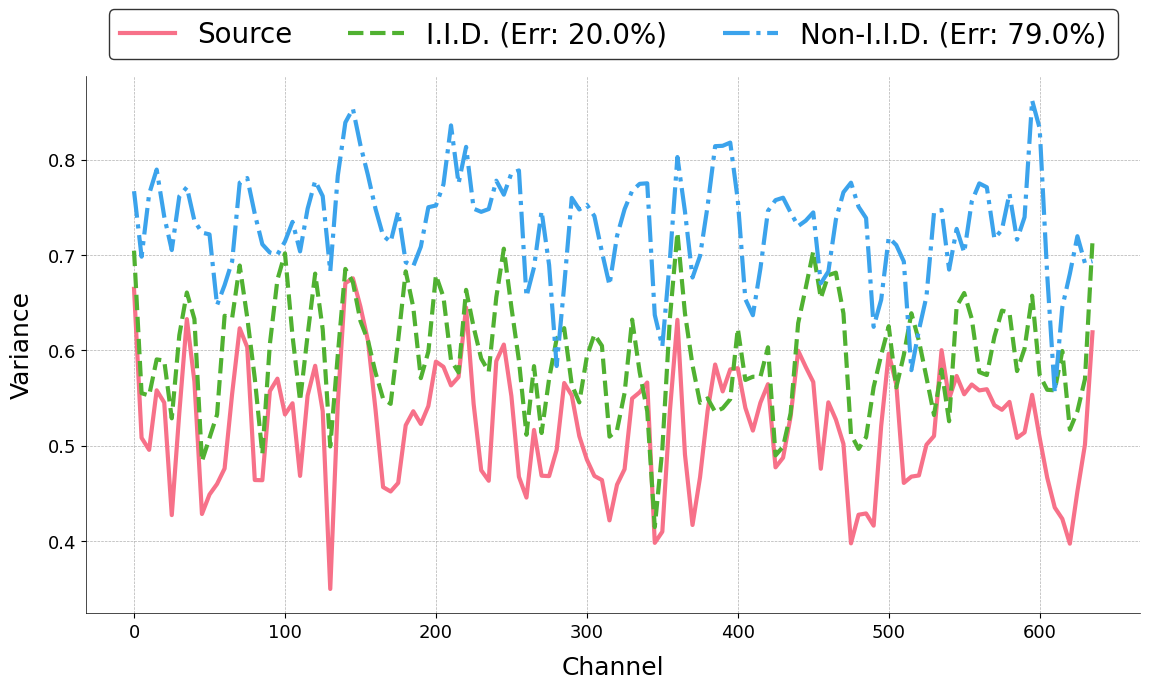}
        \caption{}
        \label{fig:phenomenon}
    \end{subfigure}
    \caption{\textbf{(a)} For clarity, we use $\bm{\mu}, \bm{\sigma}$ to denote the batch statistics, while using $\bm{m}, \bm{d}$ to denote the mean and standard deviation of the feature distribution for each sample. \textbf{(b)} Visualization of the average statistics ($\overline{\bm{d}^{2}}(\mathbf{\hat{X}}^{(i)})$) of feature distributions in a BN layer of the source model through TTBN method. We feed both source and target data streams (both i.i.d. and non-i.i.d.) and record the average variance for each channel in the corresponding BN layer. For clarity, the curves are downsampled by a factor of \num{5} and smoothed. The figure shows that the distribution statistics for the i.i.d. stream are nearing those of the source data, associated with a $20\%$ error. In contrast, the statistics for the non-i.i.d. stream notably deviate from the source, associated with a $79\%$ error.
    }
\end{figure}

\noindent \textbf{Analysis of TTBN in Non-i.i.d. Data Streams.}
\cref{fig:intro} shows the commendable efficacy of TTBN~\cite{nado2020TTBN, schneider2020TTBN} in i.i.d. test data streams, positioning it as a strong baseline method. However, its performance wanes when exposed to non-i.i.d. data streams. We attribute this degradation to misleading distribution statistics provided by non-i.i.d data batches. Specifically, TTBN’s adjustment to the source model lies on the test-time statistics\textemdash the means $\bm{\mu}$ and standard deviations (stds) $\bm{\sigma}$ of each Batch Normalization layer:
\begin{align}
    \bm{\mu} = \frac{1}{b} \sum_{i=1}^{b} \mathbf{X}^{(i)} , 
    \bm{\sigma}^2 = \frac{1}{b} \sum_{i=1}^{b} (\mathbf{X}^{(i)} - \bm{\mu})^2 , 
    \label{}
\end{align}
\noindent where $\mathbf{X}^{(i)}$ is the input feature corresponding to $i^{th}$ sample in a batch with batch size $b$.
The $\bm{\mu}$ and $\bm{\sigma}$ enable an affine transformation that normalizes the feature $\mathbf{X}^{(i)}, i \in [1, b]$ to match preferred distributions of the well-trained source model:
\begin{align}
    & \mathbf{\hat{X}}^{(i)} = \frac{\mathbf{X}^{(i)} - \bm{\mu}}{\sqrt{\bm{\sigma}^{2} + \epsilon}} = \frac{\mathbf{X}^{(i)}}{\sqrt{\bm{\sigma}^{2} + \epsilon}} + \frac{-\bm{\mu}}{\sqrt{\bm{\sigma}^{2} + \epsilon}} , \label{eq:2}\\
    &\Rightarrow\begin{cases}
        \bm{m}(\mathbf{\hat{X}}^{(i)}) &= \frac{1}{\sqrt{\bm{\sigma}^{2} + \epsilon}} \cdot \bm{m}(\mathbf{X}^{(i)}) + \frac{-\bm{\mu}}{\sqrt{\bm{\sigma}^{2} + \epsilon}} ,\\
        \bm{d}(\mathbf{\hat{X}}^{(i)}) &= \frac{1}{\sqrt{\bm{\sigma}^{2} + \epsilon}} \cdot \bm{d}(\mathbf{X}^{(i)}) ,
        \label{eq:3}
    \end{cases}
\end{align}
where $\mathbf{\hat{X}}^{(i)}$ is transformed from $\mathbf{X}^{(i)}$, and $\bm{m}(\mathbf{\hat{X}}^{(i)})$, $\bm{d}^{2}(\mathbf{\hat{X}}^{(i)})$ are the mean and variance of each transformed feature map. 
For clarity, we use $\bm{\mu}, \bm{\sigma}$ to denote the \textit{batch} statistics, while using $\bm{m}, \bm{d}$ to denote the mean and standard deviation of the feature distribution for a \textit{single} sample, as illustrated in \cref{fig:muVSm}.

\noindent \textbf{Implications for TTA.} During test-time, when fed a data batch, the TTBN layers transform their features, trying to approach $\mathbf{\hat{X}}^{(i)}$ towards the distributions of the source data features. The domain shift can be effectively mitigated and performance can be preserved if the transformed features approximate the source distribution. 
\cref{fig:phenomenon} provides a visual exploration of the impact of non-i.i.d. data streams on the transformed feature distribution. With a frozen source model, three different data streams are assessed through the TTBN method: an i.i.d. source data stream, and both i.i.d. and non-i.i.d. target data streams. These visualizations randomly spotlight one of BN layers, with the x-axis denoting feature channels, and the y-axis portraying the average variances $\overline{\bm{d}^{2}}(\mathbf{\hat{X}}^{(i)})$. The means are approximating zeros that are omitted here. 

The key takeaways include: (a) I.i.d. target data has its transformed feature distributions closer to the source distributions, and \textbf{mild performance drop} is observed (i.e., error rate = 20\%). (b) Conversely, non-i.i.d. streams manifest feature distributions that deviate from the source, correlating with the observed \textbf{significant performance dip} (i.e., error rate = 79\%). Therefore, the performance of TTBN method is greatly hampered under the non-i.i.d. (dynamic online) setting due to the drifting of target feature distributions. On the other hand, as the distribution (label) for each non-i.i.d. batch differs, it causes the gradient conflict~\cite{NEURIPS2020conflict} among batches when the model is updating towards test data. The performance is further impeded due to such conflicting optimizations. 

To this end, we aim to narrow the disparity in distributions between non-i.i.d. test features and source features by steering the feature distributions \textbf{back to the source}, ensuring they are aptly managed by the source model. Moreover, as the source distribution is set as the \textit{``reference''} for all test data streams, it sidesteps conflicting optimizations on distinctively distributed batches, thus preventing degradation of the well-trained source model.

\subsection{Distribution Alignment for TTA}

\begin{figure*}[t]
    \centering
    \includegraphics[width=0.90\textwidth]{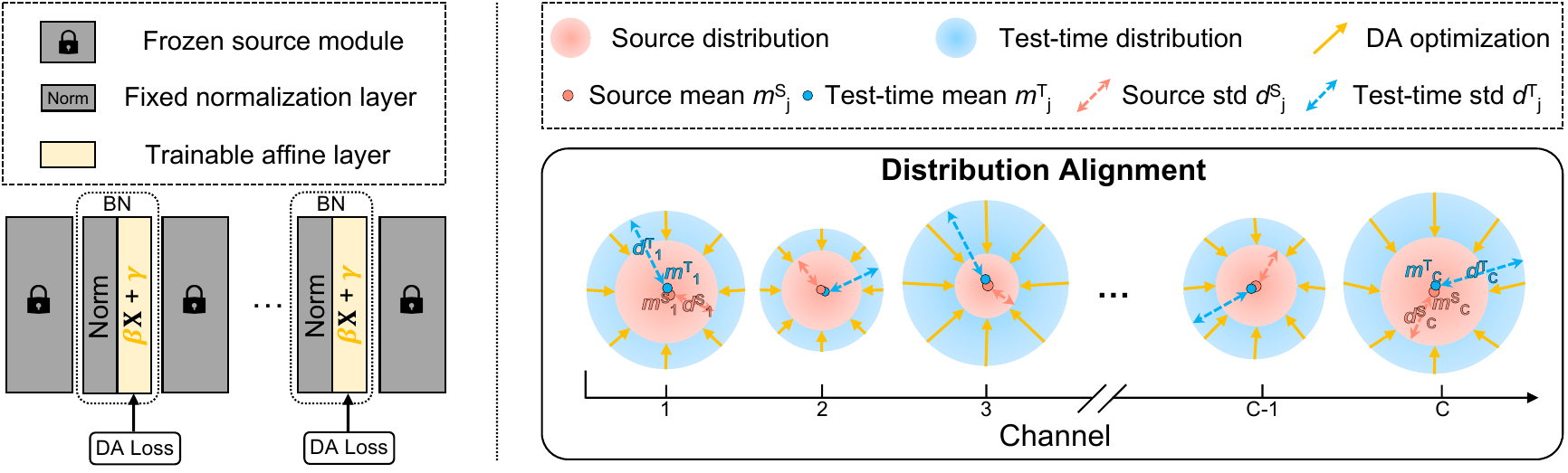}
    \caption{Illustration of the distribution alignment loss applied to features post-BN layers. This loss optimizes the affine parameters of BN layers to transform the feature distributions towards the pre-computed source distributions. Each circle represents the distribution of a channel in the feature following an affine layer, characterized by a mean (center) and a standard deviation (radius).}
    \label{fig:da}
\end{figure*}

We propose the Distribution Alignment (DA) loss, a simple yet effective method to provide consistent optimization objectives. It avoids the instability caused by conflicting objectives, and effectively counteracts domain shifts by steering the test-time feature distributions towards the source domain. The DA loss is applied to the features from intermediate layers (DA is applied to multiple layers, we omit the layer notation here for simplicity) of the source model. Upon processing a batch of input data, we calculate distribution statistics of the features in the model. For one of the intermediate features, $\mathbf{X}$, we have: 
\begin{align}
    \bm{m}_{j} = \frac{1}{H \cdot W} \sum_{p=1}^{H \cdot W} \mathbf{X}_{j,p} , \hspace{5mm} 
    {\bm{d}^2}_{j} = \frac{1}{H \cdot W} \sum_{p=1}^{H \cdot W} (\mathbf{X}_{j,p} - \bm{m}_{j})^2 , \label{eq:variance}
\end{align}
\noindent
where \( j \in \{1, \ldots, C\} \), $C$ is the channel, and $H \cdot W$ represents the spatial size of each feature map. It is important to note that prior to TTA, we pre-compute the average feature distribution statistics $\overline{\bm{m}}^S_{j}, {\overline{\bm{d}^2}}^S_{j}$ of source data offline. This computation is performed only once, after which the source statistics are retained for ongoing use. This kind of operation before deployment time is also adopted in other TTA methods, such as EATA~\cite{niu2022EATA} and RMT~\cite{dobler2023RMT}.  
During TTA, when a batch of test data is fed into the model, the test statistics $\bm{m}^T_{j}, {\bm{d}^2}^T_{j}$ are also computed using \cref{eq:variance}. Subsequently, the DA loss is computed based as: 
\begin{align}
    \mathcal{L}_{DA} =  \frac{1}{C} \sum_{j=1}^{C} \left( \left| \bm{m}^T_{j} - \overline{\bm{m}}^S_{j} \right| + \left| {\bm{d}^2}^T_{j} - {\overline{\bm{d}^2}}^S_{j} \right| \right) ,
\end{align}
\noindent
where \( \left| \cdot \right| \) denotes the absolute value operation.  Therefore, the DA loss quantifies the disparity between the feature distributions of the source and test data, with the objective of pulling test-time feature distributions back to the source domain through optimization as shown in the right side of \cref{fig:da}.

Feature distribution (means and variances) can be linearly manipulated via affine transformations. Hence, we utilize the affine layers in BN layers to be optimized by the DA loss, as depicted on the left side of \cref{fig:da}. Alternative strategies for the selection of affine layers, such as integrating external affine layers instead of utilizing those within BN layers, are discussed in Appendix G. 

At inference, BN layers utilize the population mean $\bm{\mu}_{popu}$ and variances $\bm{\sigma}_{popu}^{2}$ computed at pre-training for normalization. Before the TTA process starts, we update the such statistics in~\cite{ioffe2015BN} towards the first batch of test data as: 
\begin{align}
    \bm{\mu}_{norm} = \alpha \cdot \bm{\mu}_{popu} + (1 - \alpha) \cdot \bm{\mu}_{B_1} , \label{eq:7}\\
    \bm{\sigma}_{norm}^{2} = \alpha \cdot \bm{\sigma}_{popu}^{2} + (1 - \alpha) \cdot \bm{\sigma}_{B_1}^{2} , \label{eq:8}
\end{align}
where $\bm{\mu}_{B_1}, \bm{\sigma}_{B_1}^{2}$ are the statistics of the first batch from the test data stream, and $\alpha$ is a hyper-parameter. This modification offers a more favorable starting point for optimization, ensuring that the initial distribution discrepancy between the test and source features is not excessively large.

Additionally, we explore the synergistic effect of combining the DA loss with the entropy minimization (EM) loss:
\begin{align}
    \mathcal{L}_{EM} = \sum_{m} &\left[ \mathds{1}\left(\max_{n} \hat{y}_n > \theta\right) \sum_{n=1}^{N} -p(\hat{y}_n) \log p(\hat{y}_n) \right], \\
    &\mathcal{L}_{final} = \mathcal{L}_{DA} + \mathcal{L}_{EM},
\end{align}
where $\hat{y}_n$ denotes the predicted probability for class $n$, with $n$ ranging from 1 to $N$, $\mathds{1}(\cdot)$ denotes an indicator function, $\theta$ is the confidence threshold, and $m$ is the batch size.
\cref{sec:analyses} will reveal that, additional EM loss further improves, although DA loss alone can achieve SoTA performance.


\subsection{Domain Shift Detection in Continual TTA Setting}

In certain application scenarios, it is essential for deployed models to automatically process data streams with continual domains without manual intervention~\cite{liu2022few,chi2022metafscil}, as shown in Fig.~\ref{fig:stream}b. The DA loss is designed to pull the feature distributions in current test domain to the source distributions. When a new target domain is encountered, the model, whose affine layers are tailored to the last tested domain, may apply unsuitable affine transformations on the features of the new domain. This is particularly problematic in the event of significant domain shifts, as the discrepancy between the test-time feature distributions and the source distributions can increase substantially, thereby raising the risk of convergence to a suboptimal local minimum.

To improve the performance of our method in the continual TTA setting~\cite{song2023ecotta, dobler2023RMT, yuan2023RoTTA}, we introduce a domain shift detection mechanism. This mechanism tracks the DA loss $\mathcal{L}_{DA}$, which reflects the discrepancy between test-time feature distributions and source distributions. A domain shift is detected if the average discrepancy within a short-term window is larger than the average discrepancy within a long-term window by a predefined margin, 
$\frac{\sum_{i=0}^{p} \mathcal{L}_{DA}^{B_{t-i}}}{p} > \tau \cdot \frac{\sum_{i=0}^{q} \mathcal{L}_{DA}^{B_{t-i}}}{q}$,
where $\mathcal{L}_{DA}^{B_{t}}$ denotes the DA loss of the current batch, $p, q$ denote the lengths of short-term and long-term windows, and the $\tau$ is the threshold factor. Upon detecting a new domain, the model's trainable affine layers are reset to their initial states and the normalization layers  in BN layers are reset according to \cref{eq:7} and \cref{eq:8}. For more details on the domain shift detection mechanism, see Algorithm 1, Appendix A.
It is noteworthy that in the continual TTA setting, we employ both distribution alignment and domain shift detection, whereas for the fully TTA setting, we exclusively utilize distribution alignment.

\section{Experiments}

We conduct comparative experiments on TTA benchmarks with state-of-the-art TTA methods: TTBN~\cite{nado2020TTBN, schneider2020TTBN}, TENT~\cite{wang2021tent}, MEMO~\cite{2022MEMO}, LAME~\cite{boudiaf2022LAME}, CoTTA~\cite{wang2022CoTTA}, EATA~\cite{niu2022EATA}, NOTE~\cite{gong2022NOTE}, RoTTA~\cite{yuan2023RoTTA}, RMT~\cite{dobler2023RMT}, DELTA~\cite{zhao2023delta}, and SAR~\cite{niu2023SAR}. We also compare our method with a related SFDA work, BUFR~\cite{eastwood2022bufr}, in Appendix F.1. For a fair comparison, we adopt the codebase from RMT~\cite{dobler2023RMT} which integrates many SoTA TTA methods. In Appendix F.1, we also integrate our method into the official NOTE~\cite{gong2022NOTE} codebase for a direct comparison with NOTE~\cite{gong2022NOTE}.

\subsection{Datasets}

\noindent \textbf{CIFAR10/100-C, ImageNet-C.}
We conduct experiments on the CIFAR10-C, CIFAR100-C, and ImageNet-C~\cite{corrupted} that are common TTA benchmarks. They have \num{15} types of corruptions (target domains) applied on test and validation. Each type of corruption has \num{5} severity levels, of which we use the highest. On each target domain, CIFAR10-C/CIFAR100-C/ImageNet-C has \num{10000}/\num{10000}/\num{50000} images and \num{10}/\num{100}/\num{1000} classes.

\noindent \textbf{ImageNet-R, ImageNet-D, ImageNet-A.}
We additionally conduct experiments on other types of domain shifts. ImageNet-R~\cite{hendrycks2021imagenetr} contains \num{200} ImageNet classes with different textures and styles. ImageNet-D \cite{rusak2022imagenetd}, re-proposed from DomainNet~\cite{peng2019domainnet}, maps classes to those in ImageNet, removing unmappable classes. Furthermore, ImageNet-A~\cite{hendrycks2021imageneta} comprises adversarially filtered images from \num{200} ImageNet classes. We use these datasets as target domains, with ImageNet serving as the source domain. Appendix B shows more details.

\subsection{Implementation and Setup.}
\label{sec:experimental setup}
We evaluate our method in both fully TTA~\cite{wang2021tent, niu2022EATA, niu2023SAR, gong2022NOTE, zhao2023delta} and continual 
TTA~\cite{song2023ecotta, dobler2023RMT, yuan2023RoTTA} settings within non-i.i.d. scenarios. In the fully TTA setting, an off-the-shelf source model is online adapted to a data stream from a single target domain. In the continual TTA setting, the model is online adapted to a data stream that comprises a succession of domains, each fed sequentially one after the other.
Following most existing TTA work, we use a pre-trained WideResNet-28~\cite{widresnet}, ResNeXt-29~\cite{resxnet}, and ResNet-50~\cite{he2016deep} from the RobustBench benchmark~\cite{robustbench} as source models for the CIFAR10-C, CIFAR100-C, and ImageNet-C/D/R, respectively, in experiments for all methods. In comparative experiments, non-i.i.d. data streams of CIFAR10/100-C and ImageNet-R/D/A are generated based on Dirichlet distribution with Dirichlet parameter $\delta$ set to \num{0.1}, which controls the degree of temporal correlation of class labels in data streams. We provide illustration of non-i.i.d. data streams with different $\delta$ in Appendix C.1.
For ImageNet-C, due to the low ratio of samples per class to the number of classes, we construct non-i.i.d. data streams by sorting the images according to their labels.
More implementation details on ImageNet-D/R/A and hyper-parameter details can be found in Appendix C.2 and Appendix D.

\subsection{Main Results}

We conduct the comparative experiments in both the fully TTA (\cref{tab:sota_fully}, \cref{tab:ImageNet-DR}, \cref{tab:ImageNet-A}) and the continual TTA (\cref{tab:sota_continual}) settings as described in \cref{sec:problem}.

\begin{table*}[t]
  \centering
  \caption{Fully test-time adaptation in non-i.i.d. test data streams. Classification error rates $(\%)$ are reported for each of the \num{15} target domains within the CIFAR-10-C, CIFAR-100-C, and ImageNet-C datasets, respectively.}
  \tiny
  \begin{tabularx}{.9\textwidth}{@{}c|c|*{14}Xc|c@{}}
    \toprule
    & Method & \rule{0pt}{2.4ex}bri & gla & jpe & con & def & imp & mot & sno & zoo & fro & pix & gau & ela & sho & fog & Mean \\
    \midrule
    \midrule
    \multirow{12}{*}{\rotatebox{90}{CIFAR10-C}}
    & Source & 9.3 & 54.3 & 30.3 & 46.7 & 46.9 & 72.9 & 34.8 & 25.1 & 42.0 & 41.3 & 58.5 & 72.3 & 26.6 & 65.7 & 26.0 & 43.5 \\
    & CoTTA~\cite{wang2022CoTTA} & 75.3 & 81.5 & 78.6 & 80.3 & 77.2 & 80.1 & 77.9 & 77.4 & 77.1 & 76.4 & 77.3 & 78.3 & 78.8 & 77.3 & 76.3 & 78.0 \\
    & RMT~\cite{dobler2023RMT} & 74.3 & 81.1 & 79.6 & 75.2 & 75.9 & 80.9 & 76.0 & 77.0 & 76.0 & 76.2 & 77.5 & 79.2 & 79.1 & 78.0 & 75.6 & 77.5 \\
    & TENT~\cite{wang2021tent} & 72.5 & 81.4 & 80.1 & 76.4 & 75.2 & 81.1 & 75.1 & 76.4 & 75.8 & 75.0 & 77.2 & 78.3 & 79.0 & 78.2 & 73.6 & 77.0 \\
    & TTBN~\cite{nado2020TTBN, schneider2020TTBN} & 71.6 & 80.3 & 78.7 & 73.2 & 73.6 & 80.0 & 73.9 & 75.3 & 73.9 & 73.9 & 76.4 & 77.5 & 77.4 & 77.0 & 73.1 & 75.7\\
    & SAR~\cite{niu2023SAR} & 71.8 & 80.2 & 78.5 & 73.4 & 73.4 & 79.9 & 73.9 & 75.3 & 73.8 & 73.9 & 76.3 & 77.6 & 77.3 & 76.9 & 72.9 & 75.7 \\
    & EATA~\cite{niu2022EATA} & 75.3 & 81.5 & 78.6 & 80.3 & 77.2 & 80.1 & 77.9 & 77.4 & 77.1 & 76.4 & 77.3 & 78.3 & 78.8 & 77.3 & 76.3 & 75.7 \\
    & MEMO~\cite{2022MEMO} & 8.3 & 48.7 & 27.5 & 30.5 & 32.6 & 56.5 & 27.7 & 20.1 & 29.4 & 29.4 & 51.3 & 57.2 & 25.2 & 50.6 & 20.6 & 34.4 \\
    & LAME~\cite{boudiaf2022LAME} & \textbf{4.6} & 42.9 & \textbf{6.7} & 40.8 & 25.8 & 62.7 & \textbf{12.8} & \textbf{9.4} & \textbf{14.1} & 25.5 & 53.3 & 77.9 & \textbf{5.9} & 67.3 & \textbf{11.5} & 30.7 \\
    & NOTE~\cite{gong2022NOTE} & 13.4 & 46.3 & 44.1 & \textbf{9.9} & 24.9 & 45.3 & 20.4 & 22.6 & 25.1 & 23.5 & 36.2 & 36.9 & 36.5 & 34.3 & 21.8 & 29.4 \\
    & RoTTA~\cite{yuan2023RoTTA} & 11.0 & 45.0 & 35.3 & 21.1 & 18.8 & 46.5 & 19.6 & 23.8 & 17.2 & 23.4 & 27.3 & 37.6 & 31.6 & 35.5 & 21.1 & 27.6 \\   
    &  DA-TTA (ours) & 11.7 & \textbf{42.1} & 32.0 & 14.4 & \textbf{17.8} & \textbf{40.7} & 18.2 & 20.6 &  16.2 &  \textbf{20.6} &  \textbf{24.4} &  \textbf{31.0} &  27.6 &  \textbf{28.8} &  18.7 &  \textbf{24.3} \\
    \midrule
    \midrule
    \multirow{12}{*}{\rotatebox{90}{CIFAR100-C}}
    & Source & 29.5 & 54.1 & 41.2 & 55.1 & 29.3 & 39.4 & 30.8 & 39.5 & 28.8 & 45.8 & 74.7 & 73.0 & 37.2 & 68.0 & 50.3 &46.5 \\
    & LAME~\cite{boudiaf2022LAME} & 52.7 & 77.3 & 68.5 & 84.3 & 47.4 & 46.2 & 48.4 & 68.5 & 48.2 & 73.4 & 94.9 & 87.8 & 65.3 & 85.8 & 75.0 & 68.3 \\
    & TENT~\cite{wang2021tent} & 52.5 & 65.4 & 63.6 & 59.6 & 52.7 & 59.3 & 54.7 & 57.2 & 52.8 & 62.7 & 57.1 & 64.2 & 61.0 & 62.0 & 62.0 &59.1 \\
    & RMT~\cite{dobler2023RMT} & 50.3 & 60.1 & 59.2 & 53.7 & 51.4 & 59.1 & 53.3 & 56.1 & 51.7 & 56.1 & 53.8 & 59.7 & 56.9 & 58.7 & 57.4 & 55.8 \\
    & TTBN~\cite{nado2020TTBN, schneider2020TTBN} & 48.8 & 60.4 & 59.5 & 51.0 & 49.4 & 60.2 & 50.4 & 55.6 & 49.8 & 54.8 & 53.6 & 60.2 & 55.6 & 59.1 & 60.4 &55.3 \\
    & NOTE~\cite{gong2022NOTE} & 43.2 & 64.9 & 60.9 & 38.7 & 45.5 & 62.2 & 46.4 & 51.9 & 46.5 & 52.0 & 54.3 & 65.6 & 58.5 & 64.1 & 62.5 & 54.5 \\
    & CoTTA~\cite{wang2022CoTTA} & 48.7 & 57.8 & 55.9 & 54.1 & 50.2 & 56.9 & 51.0 & 54.8 & 49.8 & 53.7 & 51.0 & 57.2 & 54.9 & 56.3 & 60.0 & 54.2 \\
    & SAR~\cite{niu2023SAR} & 48.6 & 59.1 & 58.3 & 51.3 & 48.8 & 54.7 & 50.2 & 54.5 & 49.1 & 54.4 & 51.9 & 58.3 & 55.3 & 56.6 & 58.0 & 54.0 \\
    & EATA~\cite{niu2022EATA} & 48.6 & 58.6 & 57.8 & 51.2 & 49.2 & 53.8 & 50.3 & 53.7 & 49.7 & 54.3 & 51.9 & 57.9 & 55.6 & 56.3 & 56.8 & 53.7 \\
    & RoTTA~\cite{yuan2023RoTTA} & 30.6 & 48.1 & 48.2 & 61.9 & 32.1 & 49.9 & 34.6 & 40.6 & 32.8 & 48.2 & 40.7 & 49.2 & 41.6 & 49.2 & 45.3 & 43.5 \\
    & MEMO~\cite{2022MEMO} & 26.4 & 45.6 & 39.2 & 38.5 & 28.7 & 36.1 & 29.4 & 34.1 & 29.2 & 35.9 & 52.2 & 57.6 & 36.8 & 53.5 & 46.5 &39.3 \\    
    & DA-TTA (ours) & \textbf{23.8} & \textbf{37.8} & \textbf{37.3} & \textbf{26.6} & \textbf{25.3} & \textbf{33.5} & \textbf{27.2} & \textbf{31.5} & \textbf{25.2} & \textbf{31.4} & \textbf{29.8} & \textbf{38.5} & \textbf{32.8} & \textbf{37.3} & \textbf{36.0} & \textbf{31.6}  \\
    \midrule
    \midrule
    \multirow{12}{*}{\rotatebox{90}{ImageNet-C}}
    & Source & 41.1 & 90.2 & 68.4 & 94.6 & 82.1 & 98.2 & 85.2 & 83.1 & 77.5 & 76.7 & 79.4 & 97.8 & 83.0 & 97.1 & 75.6 & 82.0\\
    & EATA~\cite{niu2022EATA} & 95.1 & 99.7 & 98.5 & 99.6 & 99.5 & 99.5 & 99.1 & 98.5 & 98.5 & 98.6 & 98.5 & 99.5 & 98.2 & 99.4 & 97.9 & 98.7 \\
    & TENT~\cite{wang2021tent} & 92.7 & 98.5 & 95.5 & 99.2 & 98.2 & 97.8 & 98.0 & 97.0 & 96.0 & 97.0 & 95.1 & 97.8 & 96.0 & 97.8 & 94.2 & 96.7 \\
    & RMT~\cite{dobler2023RMT} & 93.5 & 98.3 & 95.7 & 97.6 & 98.1 & 97.4 & 96.9 & 96.1 & 96.2 & 96.1 & 95.2 & 97.4 & 95.9 & 97.6 & 94.6 & 96.4 \\
    & SAR~\cite{niu2023SAR} & 91.9 & 98.2 & 94.6 & 98.9 & 98.1 & 97.6 & 97.8 & 96.3 & 95.7 & 96.7 & 94.6 & 97.8 & 95.3 & 97.9 & 93.6 & 96.3 \\
    & CoTTA~\cite{wang2022CoTTA} & 91.8 & 98.2 & 94.9 & 98.0 & 98.1 & 97.2 & 96.7 & 95.3 & 95.6 & 95.0 & 94.4 & 97.3 & 95.2 & 97.1 & 93.7 & 95.9 \\
    & TTBN~\cite{nado2020TTBN, schneider2020TTBN} & 91.1 & 98.1 & 94.8 & 97.4 & 98.0 & 97.5 & 96.6 & 95.1 & 95.4 & 94.8 & 94.3 & 97.5 & 95.1 & 97.4 & 93.6 & 95.8 \\
    & NOTE~\cite{gong2022NOTE} & 66.0 & 95.7 & 91.4 & 93.5 & 97.1 & 95.6 & 89.5 & 85.3 & 89.3 & 84.6 & 90.5 & 96.0 & 83.3 & 96.1 & 79.5 & 88.9\\
    & DELTA~\cite{zhao2023delta} & 50.9 & 90.2 & 69.2 & 97.5 & 90.4 & 86.5 & 87.2 & 78.7 & 74.2 & 78.2 & 64.7 & 86.6 & 68.2 & 85.8 & 64.0 & 78.1\\
    & MEMO~\cite{2022MEMO} & 39.6 & 86.4 & 63.8 & 87.2 & 81.4 & 91.3 & 77.5 & 71.7 & 72.5 & 69.5 & 65.0 & 92.5 & 72.8 & 91.3 & 65.4 & 75.2 \\
    & RoTTA~\cite{yuan2023RoTTA} & 37.8 & 86.6 & 63.2 & 83.2 & 86.4 & 84.8 & 76.6 & 67.9 & 64.7 & 73.3 & 55.5 & 85.8 & 58.3 & 85.7 & 54.6 & 71.0 \\
    & LAME~\cite{boudiaf2022LAME} & \textbf{27.6} & 83.3 & \textbf{45.6} & 90.2 & \textbf{57.3} & 98.9 & 71.6 & 75.7 & \textbf{60.2} & \textbf{52.6} & 62.6 & 98.8 & 78.4 & 97.8 & 60.5 & 70.7 \\    
    &  DA-TTA (ours) & 35.2 & \textbf{77.3} & 55.2 & \textbf{70.8} & 75.5 & \textbf{78.0} & \textbf{69.5} & \textbf{63.9} & 60.8 & 74.9 & \textbf{50.3} & \textbf{79.7} & \textbf{53.4} & \textbf{78.6} & \textbf{49.7} & \textbf{64.8} \\
    \bottomrule
  \end{tabularx}
  \label{tab:sota_fully}
\end{table*}

\noindent \textbf{Fully TTA in Non-I.I.D. Data Streams.}
\cref{tab:sota_fully} presents the results of our method in comparison to other TTA methods on the commonly used corruption benchmark. Observing the ``Mean'' column reveals that over half of the prior methods yield results inferior to the source model without adaptation, suggesting an adaptation failure. Our method, denoted DA-TTA, outperforms competing methods across all the datasets, showcasing accuracy improvements of $3.3\%, 7.7\%, 5.9\%$ over the next best-performing methods, respectively.
Furthermore, DA-TTA demonstrates robust performance across all target domains. In contrast, LAME excels in certain domains but significantly underperforms or even regresses relative to the ``Source'' in others. 
Besides, it is observed that many previous methods exhibit notably poorer results in non-i.i.d. streams compared to i.i.d. streams (shown in Appendix F.2). Our method, however, obtains close results in the i.i.d. and non-i.i.d. data streams.

\begin{table}[t]
\centering
\begin{minipage}[t]{0.63\linewidth}
\centering
\scriptsize
\caption{Fully test-time adaptation in non-i.i.d. streams. Presented are error rates $(\%)$ on ImageNet-D and ImageNet-R datasets.}
\begin{tabular}{@{}c|*{5}c|c|c@{}}
    \toprule
     \multirow{2}{*}{Method} & \multicolumn{6}{c|}{ImageNet-D}  & \multirow{2}{*}{ImageNet-R} \\
     \cline{2-7} 
     & \raisebox{-2pt}{cli} & \raisebox{-2pt}{inf} & \raisebox{-2pt}{pai} & \raisebox{-2pt}{real} & \raisebox{-2pt}{ske} & \raisebox{-2pt}{Mean}\\
    \midrule
    \midrule
    Source & 76.0 & 89.7 & 65.1 & 40.2 & 82.0 & 70.6 & 63.8  \\
    LAME~\cite{boudiaf2022LAME} & 88.0 & 99.7 & 70.5 & \textbf{35.4} & 92.6 & 77.2 & 84.8  \\
    TTBN~\cite{nado2020TTBN, schneider2020TTBN}  & 81.4 & 91.3 & 79.0 & 74.5 & 87.3 & 82.7 & 69.2 \\
    TENT~\cite{wang2021tent} & 80.9 & 90.8 & 79.2 & 76.6 & 87.6 & 83.0 & 68.3  \\
    CoTTA~\cite{song2023ecotta} & 80.0 & 90.6 & 77.5 & 73.0 & 85.8 & 81.4  & 67.9 \\
    SAR~\cite{niu2023SAR} & 80.3 & 90.0 & 78.5 & 74.9 & 86.0 & 81.9 & 67.6  \\
    EATA~\cite{niu2022EATA} & 79.6 & 89.4 & 78.1 & 75.5 & 86.2 & 81.8  & 66.3  \\
    RoTTA~\cite{yuan2023RoTTA}& 70.3 & 86.0 & 63.3 & 40.7 & 77.6 & 67.6  & 61.5  \\
    DA-TTA(ours) & \textbf{69.2} & \textbf{85.4} & \textbf{61.5} & 40.7 & \textbf{75.9} & \textbf{66.5} & \textbf{58.5}  \\
    \bottomrule
  \end{tabular}
  \label{tab:ImageNet-DR}
\end{minipage}
\hfill
\begin{minipage}[t]{0.35\linewidth}
\centering
\scriptsize
\caption{Fully test-time adaptation. Error rates $(\%)$ on ImageNet-A dataset.}
\begin{tabular}{@{}c|c|c@{}}
    \toprule
     Method & i.i.d. & non-i.i.d. \\
    \midrule
    \midrule
    Source & 90.5 & 90.5 \\
    LAME~\cite{boudiaf2022LAME} & 95.2 & 95.3 \\
    TTBN~\cite{nado2020TTBN, schneider2020TTBN}  & 92.8 & 93.0 \\
    RMT~\cite{dobler2023RMT}  & 95.5 & 95.5 \\
    TENT~\cite{wang2021tent} & 92.5 & 92.6 \\
    CoTTA~\cite{song2023ecotta} & 93.0 & 93.2\\
    SAR~\cite{niu2023SAR} & 93.5 & 93.7\\
    EATA~\cite{niu2022EATA} & 92.8 & 92.7\\
    RoTTA~\cite{yuan2023RoTTA}& 93.5 & 93.1 \\ 
    DA-TTA(ours) & \textbf{86.8} & \textbf{88.6} \\
    \bottomrule
  \end{tabular}
  \label{tab:ImageNet-A}
\end{minipage}
\end{table}

Apart from the corruption domain shifts, \cref{tab:ImageNet-DR} presents the results of evaluations on realistic domain shifts using the ImageNet-D and ImageNet-R datasets. Previous methods, except for RoTTA, fail to adapt in these non-i.i.d. streams (performing worse than the ``Source''), while our method achieves an improvement of $3.9\%$ on ImageNet-D and $5.3\%$ on ImageNet-R compared to the ``Source''. Additionally, \cref{tab:ImageNet-A} shows the results on the ImageNet-A dataset. Our method demonstrates effectiveness in adapting to the adversarial attack domain shift.

\begin{table*}[t]
  \centering
  \caption{Continual test-time adaptation in non-i.i.d. test data stream. Presented are the classification error rates $(\%)$ for TTA methods that are continually adapting to \num{15} target domains within the ImageNet-C dataset under a non-i.i.d. data stream scenario.}
  \tiny
  \begin{tabularx}{.9\textwidth}{c|*{14}Xc|c}
    \toprule
    \multirow{2}{*}{\raisebox{-0.22\normalbaselineskip}{Method}} & \multicolumn{15}{c|}{\textbf{\raisebox{0.19\normalbaselineskip}{\rotatebox{90}{\rule{0.35pt}{1cm}}} Continual adaptation to sequential domains $\xrightarrow[\hspace{1cm}]{}$ }}
    & \multirow{2}{*}{\raisebox{-0.22\normalbaselineskip}{Mean}}\\
    \cline{2-16}
    & \rule{0pt}{2.4ex}bri & gla & jpe & con & def & imp & mot & sno & zoo & fro & pix & gau & ela & sho & fog & \\
    \midrule
    \midrule
    Source & 41.1 & 90.2 & 68.4 & 94.6 & 82.1 & 98.2 & 85.2 & 83.1 & 77.5 & 76.7 & 79.4 & 97.8 & 83.0 & 97.1 & 75.6 & 82.0 \\
    EATA~\cite{niu2022EATA} & 95.1 & 99.5 & 99.5 & 99.8 & 99.8 & 99.8 & 99.8 & 99.7 & 99.8 & 99.8 & 99.8 & 99.9 & 99.8 & 99.8 & 99.8 & 99.4 \\
    TENT~\cite{wang2021tent} & 92.7 & 98.9 & 99.1 & 99.8 & 99.8 & 99.8 & 99.8 & 99.8 & 99.8 & 99.8 & 99.8 & 99.8 & 99.8 & 99.8 & 99.8 & 99.2 \\
    SAR~\cite{niu2023SAR} & 92.0 & 98.7 & 97.8 & 99.6 & 99.7 & 99.7 & 99.8 & 99.8 & 99.8 & 99.8 & 99.8 & 99.8 & 99.8 & 99.8 & 99.8 & 99.0 \\
    CoTTA~\cite{wang2022CoTTA} & 91.8 & 98.0 & 96.4 & 98.2 & 98.7 & 98.7 & 99.0 & 99.1 & 99.2 & 99.4 & 99.3 & 99.4 & 99.5 & 99.5 & 99.6 & 98.4 \\
    RMT~\cite{dobler2023RMT} & 93.6 & 98.2 & 96.5 & 98.2 & 98.5 & 98.2 & 98.5 & 98.3 & 98.5 & 98.7 & 97.7 & 98.5 & 97.7 & 98.7 & 97.6 & 97.8 \\
    TTBN~\cite{nado2020TTBN, schneider2020TTBN} & 91.1 & 98.1 & 94.8 & 97.4 & 98.0 & 97.5 & 96.6 & 95.1 & 95.4 & 94.8 & 94.3 & 97.5 & 95.1 & 97.4 & 93.6 & 95.8 \\
    DELTA~\cite{zhao2023delta} & 50.9 & 90.2 & 78.9 & 96.6 & 94.1 & 93.2 & 93.6 & 92.5 & 91.1 & 94.4 & 90.8 & 95.7 & 93.5 & 95.7 & 95.2 & 89.8 \\
    RoTTA~\cite{yuan2023RoTTA} & 37.7 & 86.6 & 65.0 & 96.1 & 85.7 & 89.9 & 88.4 & 84.7 & 86.2 & 87.0 & 84.0 & 94.1 & 87.4 & 93.0 & 96.6 & 84.2 \\
    LAME~\cite{boudiaf2022LAME} & \textbf{27.6} & \textbf{83.3} & \textbf{45.6} & 90.2 & \textbf{57.3} & 98.9 & \textbf{71.6} & 75.7 & \textbf{60.1} & \textbf{52.6} & 62.6 & 98.8 & 78.4 & 97.7 & 60.4 & 70.7 \\
    DA-TTA (ours) & 36.9 & 88.9 & 59.4 & \textbf{78.4} & 78.3 & \textbf{82.5} & 74.1 & \textbf{67.2} & 63.8 & 73.4 & \textbf{55.4} & \textbf{81.8} & \textbf{56.9} & \textbf{79.3} & \textbf{52.1} & \textbf{68.6} \\
    \bottomrule
  \end{tabularx}
  \label{tab:sota_continual}
\end{table*} 
\noindent \textbf{Continual TTA in Non-I.I.D. Data Streams.}
In \cref{tab:sota_continual}, we present the comparison of our method with other TTA methods when applied to a test data stream comprised of continual domains. This demanding stream delivers a sequence of \num{750000} images from \num{15} target domains of ImageNet-C under the non-i.i.d. sampling condition.
It is observed that the majority of competing methods present incur error rates in excess of $90\%$, significantly underperforming when compared to the ``Source''. While LAME performs optimally on a few target domains, it encounters failures in several domains, registering error rates above $90\%$. In comparison,  DA-TTA showcases robust adaptation capabilities on all target domains and achieves the best overall average performance.

\begin{figure}[h]
    \centering
    
    \begin{subfigure}{0.35\textwidth}
        \includegraphics[width=\linewidth]{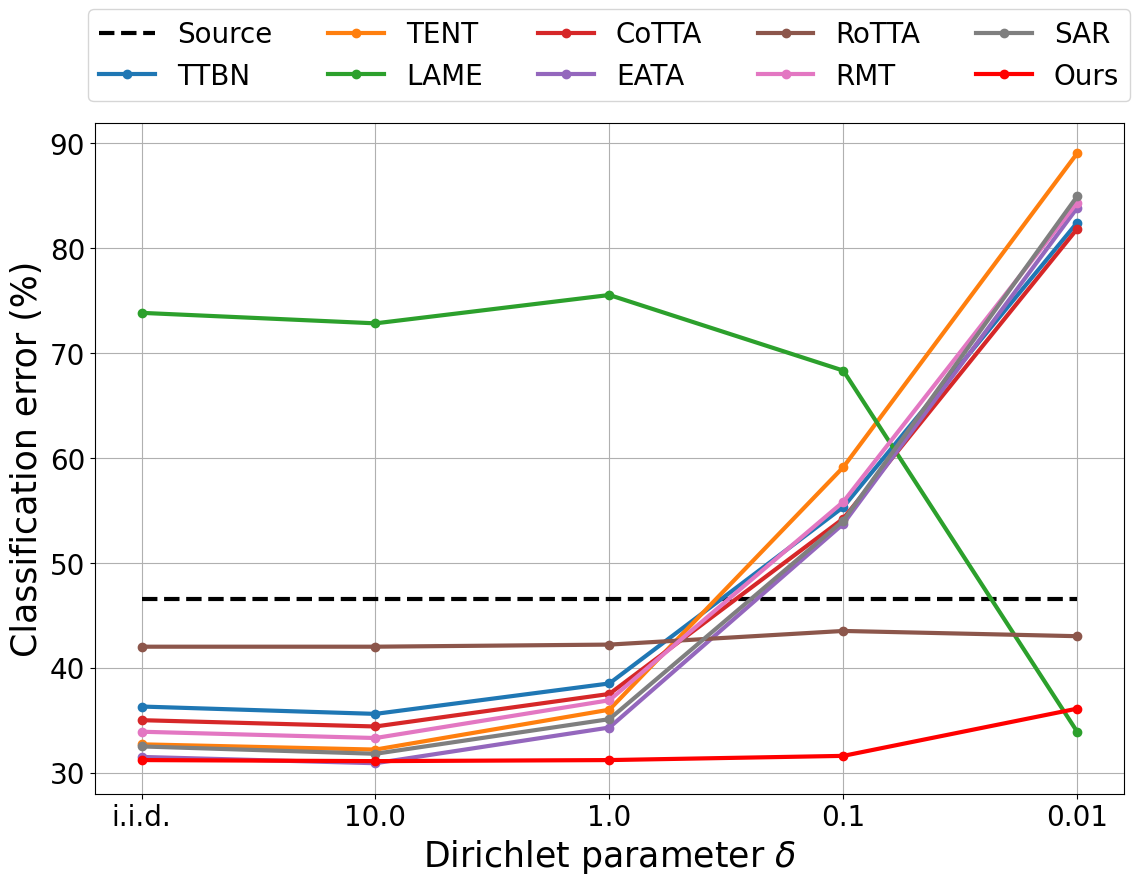}
        \caption{Effect of Dirichlet parameter.}
        \label{fig:dirichlet}
    \end{subfigure}
    \hspace{0.5cm}
    \begin{subfigure}{0.35\textwidth}
        \includegraphics[width=\linewidth]{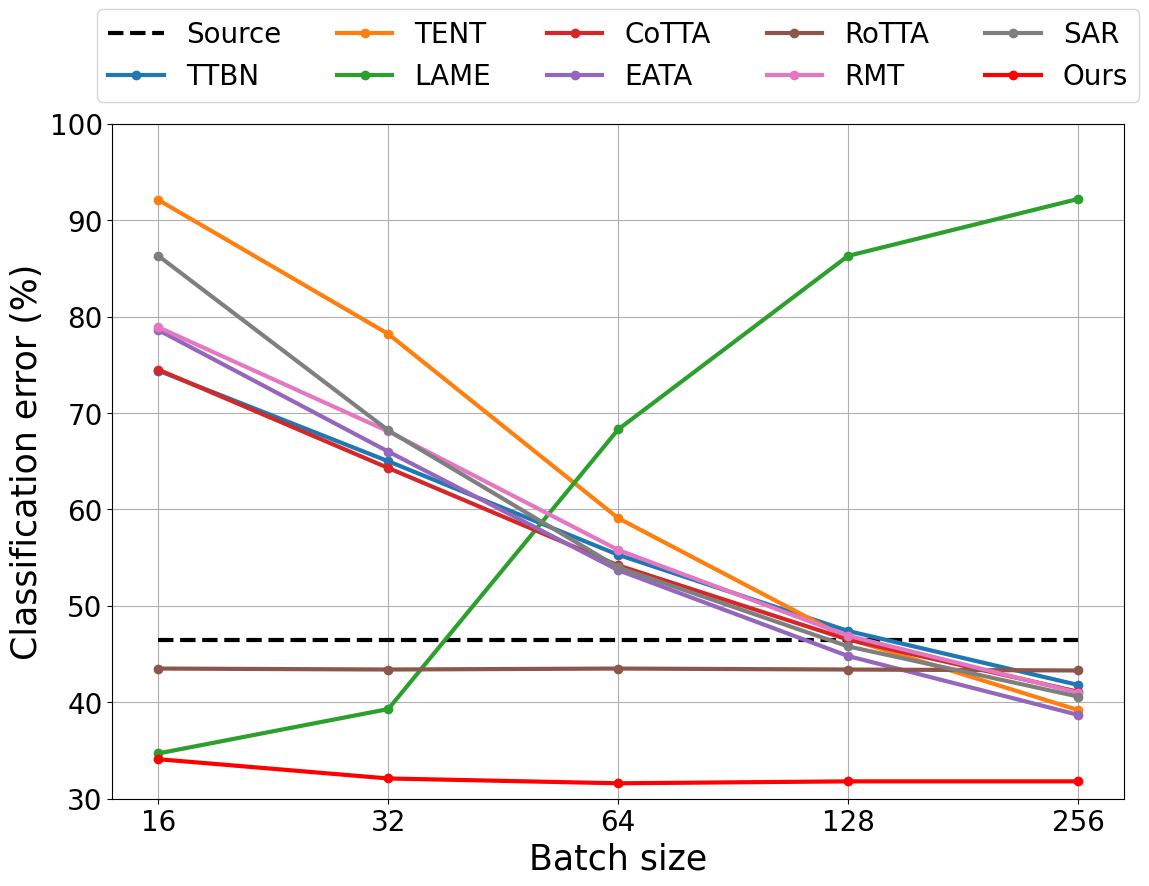}
        \caption{Effect of batch size. }
        \label{fig:batch_size}
    \end{subfigure}
    \caption{Robustness on various conditions of test data stream.}
    \label{fig:dirichlet_and_batch_size}
\end{figure}
\noindent \textbf{Robustness on Different Conditions of Data Streams.}
We examine the effect of the non-i.i.d. degree and the batch size. As illustrated in \cref{fig:dirichlet}, a smaller Dirichlet parameter $\delta$ indicates a higher degree of temporal correlation within the data stream. Most existing TTA methods experience a marked performance decline as $\delta$ decreases. LAME excels under intense temporal correlation, yet it underperforms compared to the baseline in less severe cases. In contrast, our method maintains robust performance across various degrees of non-i.i.d. severity.
\cref{fig:batch_size} illustrates the impact of varying batch sizes. It's observed that most existing TTA methods experience a decline in performance as the batch size is reduced. This phenomenon could be attributed to larger batches more accurately representing the target domain's distribution, thereby reducing the conflict with the optimization objective. In contrast, our method shows consistent performance, proving to be robust across different batch sizes.

\subsection{Ablation Studies}
\label{sec:analyses}

\noindent \textbf{Effects of Model Components.}
As detailed in \cref{tab:ablation}, we conduct an ablation study in non-i.i.d. data streams across three datasets. Firstly, we explored the effects of applying DA optimization within different ranges of the source model. The terms `\textit{w/o} low-level DA' and `\textit{w/o} high-level DA' refer to the application of DA optimization to the latter and former halves of the affine layers in the model, respectively. The results indicate a performance decline when the DA optimization range is reduced. However, the decrease in performance is relatively modest. This can be attributed to the correlated nature of feature distributions among layers within the frozen model, ensuring distributions in layers not directly supervised remain controlled. Moreover, the application of EM on top of the TTBN baseline, which is the TENT method, yields diminished results in non-i.i.d. data streams, as shown in \cref{tab:sota_fully}. Nevertheless, introducing an EM loss atop the DA loss resulted in enhanced performance, highlighting the synergistic effect of the EM loss under protection from DA optimization.

\begin{table}
  \centering
  \scriptsize
  \caption{Ablation study in error rates. `\textit{w/o} low-level DA' and `\textit{w/o} high-level DA' denote removing the DA supervision from the first and second halves of the model. 
}
  \begin{tabular}{@{}c|ccc@{}}
    \toprule
    Method & CIFAR10-C & CIFAR100-C & ImageNet-C  \\
    \midrule
    \midrule
    Source & 43.5 & 46.5 & 82.0 \\
    \textit{w/o} low-level DA & 26.6 & 32.6 & 68.1  \\
    \textit{w/o} high-level DA & 24.9 & 33.5 & 67.1  \\
    \textit{w/o} $\mathcal{L}_{EM}$ & 28.1 & 35.8 & 70.1  \\
    Ours & \textbf{24.3} & \textbf{31.6} & \textbf{64.8}  \\
    \bottomrule
  \end{tabular}
  \label{tab:ablation}
\end{table}

\begin{figure*}[h]
    \centering
    \begin{subfigure}{0.5\textwidth}
        \includegraphics[width=\linewidth]{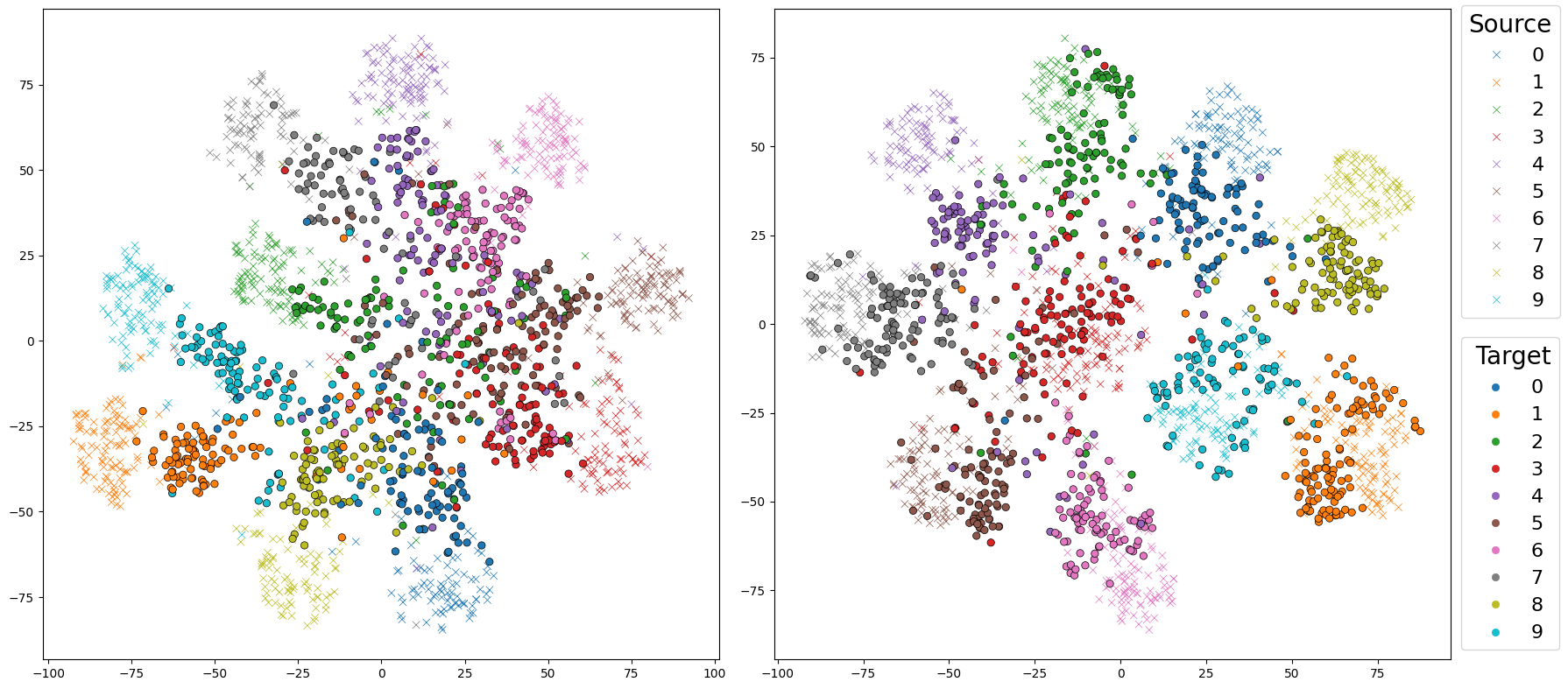}
        \caption{}
        \label{fig:tsne}
    \end{subfigure}
    \hspace{0.5cm}
    \begin{subfigure}{0.4\textwidth}
        \includegraphics[width=\linewidth]{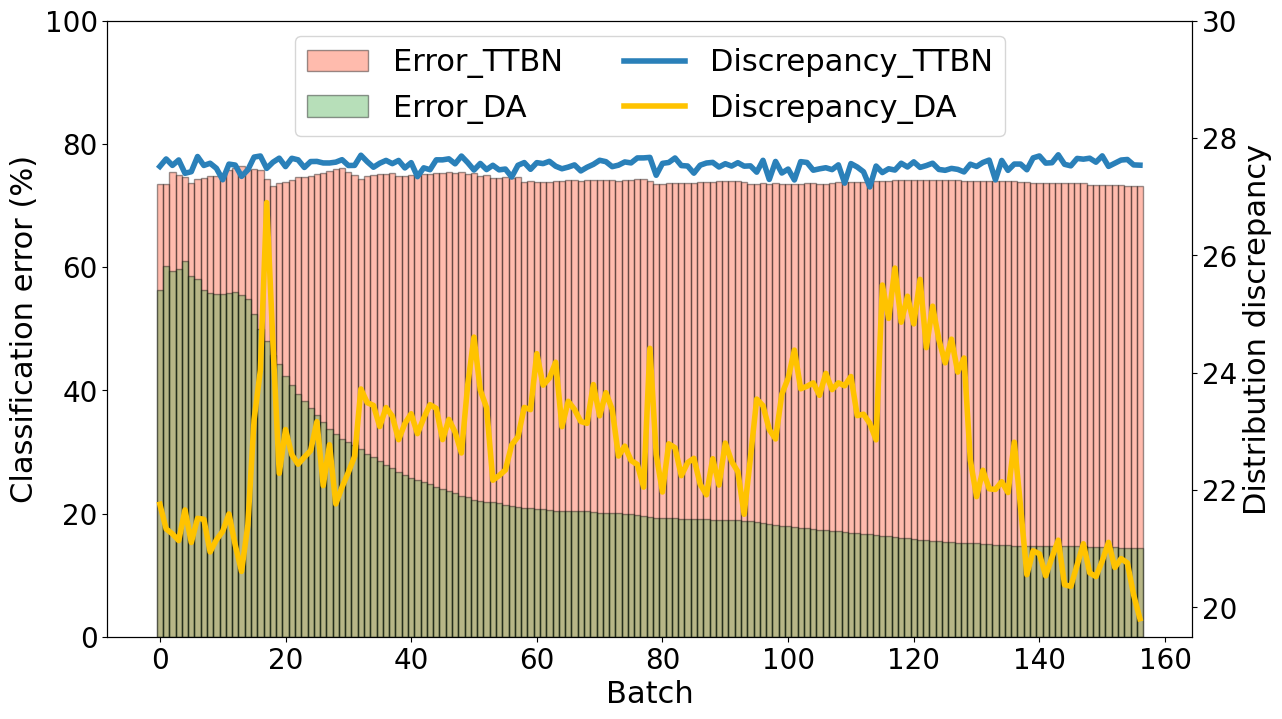}
        \caption{}
        \label{fig:objective_alignment}
    \end{subfigure}
    \caption{(a) DA alleviates domain shifts. Visualization of features prior to the classifier of the model for both source data and target data (one of target domains in CIFAR10-C dataset). Left: Source/target data inputs source model. Right: Source data inputs source model, and target data inputs source model with DA. (b) Verifying alignment between DA optimization and task objective from correlation between distribution discrepancy and classification error.}
\end{figure*}
\noindent \textbf{DA Alleviates Domain Shifts.}
In this analysis, we input source data and target domain data into the source model, labeling the features prior to the classifier layer as $\mathbf{R}_S$ for the source and $\mathbf{R}_T$ for the target, respectively. We then feed target data under our TTA method, labeling these features as $\mathbf{R'}_T$.
Visualization of $\mathbf{R}_S$ and $\mathbf{R}_T$ is provided in the left plot in \cref{fig:tsne} via t-SNE, while the right plot depicts the $\mathbf{R}_S$ and $\mathbf{R'}_T$. Upon examining the transition from the left to the right, we observe a clear trend: the features of each class in the target domain not only become discriminative but also show an alignment with their corresponding classes in the source domain. This convergence of class-specific clusters confirms that our method is successfully reducing domain shift by steering the target feature distributions back to those from the source.

\noindent \textbf{Alignment Between DA Optimization and Task Objective.}
DA optimization minimizes the discrepancy in distribution between test-time and source features. The task objective is to classify the incoming data stream. \cref{fig:objective_alignment} provides a visualization of both the distribution discrepancy and the cumulative classification errors across data batches. Notably, there is a trend of decreasing accumulated error, which corresponds with the shrunken distribution discrepancy, in contrast to the larger discrepancy observed in the TTBN baseline.

\section{Conclusion}
In this paper, we introduce a simple yet effective Distribution Alignment (DA) method for fully realizing test-time adaptation within dynamic online streams. Our proposed distribution alignment loss aligns test-time data features with the source distributions, ensuring compatibility with the source model and addressing the challenges posed by label shifts across online data batches. The addition of a domain shift detection mechanism further strengthens our method's performance in environments with continual domain shifts. Extensive experiments confirm the superiority of our method in non-i.i.d. streams, while it also maintains competitive performance under the i.i.d. assumption.

\section*{Acknowledgements}
This work is supported by an NSERC Discovery grant, the Gina Cody Research and Innovation Fellowship, and in part by the National Natural Science Foundation of China under Grant 62171269.

%
%
\bibliographystyle{splncs04}
\bibliography{main}

\begin{thebibliography}{10}
\providecommand{\url}[1]{\texttt{#1}}
\providecommand{\urlprefix}{URL }
\providecommand{\doi}[1]{https://doi.org/#1}

\bibitem{ahmed2023ssda}
Ahmed, S., Al~Arafat, A., Rizve, M.N., Hossain, R., Guo, Z., Rakin, A.S.: Ssda: Secure source-free domain adaptation. In: ICCV. pp. 19180--19190 (2023)

\bibitem{bartler2022mt3}
Bartler, A., B{\"u}hler, A., Wiewel, F., D{\"o}bler, M., Yang, B.: Mt3: Meta test-time training for self-supervised test-time adaption. In: International Conference on Artificial Intelligence and Statistics. pp. 3080--3090 (2022)

\bibitem{boudiaf2022LAME}
Boudiaf, M., Mueller, R., Ben~Ayed, I., Bertinetto, L.: Parameter-free online test-time adaptation. In: CVPR. pp. 8344--8353 (2022)

\bibitem{chen2018domain}
Chen, Y., Li, W., Sakaridis, C., Dai, D., Van~Gool, L.: Domain adaptive faster r-cnn for object detection in the wild. In: CVPR. pp. 3339--3348 (2018)

\bibitem{chi2022metafscil}
Chi, Z., Gu, L., Liu, H., Wang, Y., Yu, Y., Tang, J.: Metafscil: A meta-learning approach for few-shot class incremental learning. In: IEEE Conference on Computer Vision and Pattern Recognition (2022)

\bibitem{chiadapting}
Chi, Z., Gu, L., Zhong, T., Liu, H., YU, Y., Plataniotis, K.N., Wang, Y.: Adapting to distribution shift by visual domain prompt generation. In: International Conference on Learning Representations

\bibitem{chi2020all}
Chi, Z., Mohammadi~Nasiri, R., Liu, Z., Lu, J., Tang, J., Plataniotis, K.N.: All at once: Temporally adaptive multi-frame interpolation with advanced motion modeling. In: European Confererence on Computer Vison (2020)

\bibitem{chi2021test}
Chi, Z., Wang, Y., Yu, Y., Tang, J.: Test-time fast adaptation for dynamic scene deblurring via meta-auxiliary learning. In: IEEE Conference on Computer Vision and Pattern Recognition (2021)

\bibitem{robustbench}
Croce, F., Andriushchenko, M., Sehwag, V., Debenedetti, E., Flammarion, N., Chiang, M., Mittal, P., Hein, M.: Robustbench: a standardized adversarial robustness benchmark. In: NeurIPS (2021)

\bibitem{ding2022source}
Ding, N., Xu, Y., Tang, Y., Xu, C., Wang, Y., Tao, D.: Source-free domain adaptation via distribution estimation. In: CVPR. pp. 7212--7222 (2022)

\bibitem{dobler2023RMT}
D{\"o}bler, M., Marsden, R.A., Yang, B.: Robust mean teacher for continual and gradual test-time adaptation. In: CVPR. pp. 7704--7714 (2023)

\bibitem{eastwood2022bufr}
Eastwood, C., Mason, I., Williams, C., Sch{\"o}lkopf, B.: Source-free adaptation to measurement shift via bottom-up feature restoration. In: ICLR (2022)

\bibitem{fernando2013unsupervised}
Fernando, B., Habrard, A., Sebban, M., Tuytelaars, T.: Unsupervised visual domain adaptation using subspace alignment. In: ICCV. pp. 2960--2967 (2013)

\bibitem{ganin2015unsupervised}
Ganin, Y., Lempitsky, V.: Unsupervised domain adaptation by backpropagation. In: ICML. pp. 1180--1189 (2015)

\bibitem{gao2023dda}
Gao, J., Zhang, J., Liu, X., Darrell, T., Shelhamer, E., Wang, D.: Back to the source: Diffusion-driven adaptation to test-time corruption. In: CVPR. pp. 11786--11796 (2023)

\bibitem{gong2022NOTE}
Gong, T., Jeong, J., Kim, T., Kim, Y., Shin, J., Lee, S.J.: Note: Robust continual test-time adaptation against temporal correlation. In: NeurIPS. vol.~35, pp. 27253--27266 (2022)

\bibitem{Shayan2024Fed}
Hamidi, S.M., Tan, R., Ye, L., Yang, E.H.: Fed-it: Addressing class imbalance in federated learning through an information-theoretic lens. In: 2024 IEEE International Symposium on Information Theory (ISIT) (2024)

\bibitem{he2016deep}
He, K., Zhang, X., Ren, S., Sun, J.: Deep residual learning for image recognition. In: CVPR. pp. 770--778 (2016)

\bibitem{hendrycks2021imagenetr}
Hendrycks, D., Basart, S., Mu, N., Kadavath, S., Wang, F., Dorundo, E., Desai, R., Zhu, T., Parajuli, S., Guo, M., Song, D., Steinhardt, J., Gilmer, J.: The many faces of robustness: A critical analysis of out-of-distribution generalization. ICCV  (2021)

\bibitem{corrupted}
Hendrycks, D., Dietterich, T.: Benchmarking neural network robustness to common corruptions and perturbations. In: ICLR (2019)

\bibitem{hendrycks2021imageneta}
Hendrycks, D., Zhao, K., Basart, S., Steinhardt, J., Song, D.: Natural adversarial examples. CVPR  (2021)

\bibitem{huang2021model}
Huang, J., Guan, D., Xiao, A., Lu, S.: Model adaptation: Historical contrastive learning for unsupervised domain adaptation without source data. In: NeurIPS. vol.~34, pp. 3635--3649 (2021)

\bibitem{ioffe2015BN}
Ioffe, S., Szegedy, C.: Batch normalization: Accelerating deep network training by reducing internal covariate shift. In: ICML. pp. 448--456 (2015)

\bibitem{kang2019contrastive}
Kang, G., Jiang, L., Yang, Y., Hauptmann, A.G.: Contrastive adaptation network for unsupervised domain adaptation. In: CVPR. pp. 4893--4902 (2019)

\bibitem{koh2021wilds}
Koh, P.W., Sagawa, S., Marklund, H., Xie, S.M., Zhang, M., Balsubramani, A., Hu, W., Yasunaga, M., Phillips, R.L., Gao, I., et~al.: Wilds: A benchmark of in-the-wild distribution shifts. In: ICML. pp. 5637--5664 (2021)

\bibitem{kundu2022balancing}
Kundu, J.N., Kulkarni, A.R., Bhambri, S., Mehta, D., Kulkarni, S.A., Jampani, V., Radhakrishnan, V.B.: Balancing discriminability and transferability for source-free domain adaptation. In: ICML. pp. 11710--11728 (2022)

\bibitem{kurmi2021domain}
Kurmi, V.K., Subramanian, V.K., Namboodiri, V.P.: Domain impression: A source data free domain adaptation method. In: Proceedings of the IEEE/CVF winter conference on applications of computer vision. pp. 615--625 (2021)

\bibitem{li2021transferable}
Li, S., Xie, M., Gong, K., Liu, C.H., Wang, Y., Li, W.: Transferable semantic augmentation for domain adaptation. In: CVPR. pp. 11516--11525 (2021)

\bibitem{liang2020we}
Liang, J., Hu, D., Feng, J.: Do we really need to access the source data? source hypothesis transfer for unsupervised domain adaptation. In: ICML. pp. 6028--6039 (2020)

\bibitem{lim2023ttn}
Lim, H., Kim, B., Choo, J., Choi, S.: {TTN}: A domain-shift aware batch normalization in test-time adaptation. In: ICLR (2023)

\bibitem{lin2022prototype}
Lin, H., Zhang, Y., Qiu, Z., Niu, S., Gan, C., Liu, Y., Tan, M.: Prototype-guided continual adaptation for class-incremental unsupervised domain adaptation. In: ECCV. pp. 351--368 (2022)

\bibitem{liu2023meta}
Liu, H., Chi, Z., Yu, Y., Wang, Y., Chen, J., Tang, J.: Meta-auxiliary learning for future depth prediction in videos. In: IEEE Winter Conference on Applications of Computer Vision (2023)

\bibitem{liu2022few}
Liu, H., Gu, L., Chi, Z., Wang, Y., Yu, Y., Chen, J., Tang, J.: Few-shot class-incremental learning via entropy-regularized data-free replay. In: European Conference on Computer Vision (2022)

\bibitem{liu2021ttt++}
Liu, Y., Kothari, P., Van~Delft, B., Bellot-Gurlet, B., Mordan, T., Alahi, A.: Ttt++: When does self-supervised test-time training fail or thrive? In: NeurIPS. vol.~34, pp. 21808--21820 (2021)

\bibitem{long2015learning}
Long, M., Cao, Y., Wang, J., Jordan, M.: Learning transferable features with deep adaptation networks. In: ICML. pp. 97--105 (2015)

\bibitem{long2018conditional}
Long, M., Cao, Z., Wang, J., Jordan, M.I.: Conditional adversarial domain adaptation. In: NeurIPS. vol.~31 (2018)

\bibitem{marsden2022gtta}
Marsden, R.A., D{\"o}bler, M., Yang, B.: Introducing intermediate domains for effective self-training during test-time. arXiv preprint arXiv:2208.07736  (2022)

\bibitem{marsden2023boid}
Marsden, R.A., D{\"o}bler, M., Yang, B.: Universal test-time adaptation through weight ensembling, diversity weighting, and prior correction. arXiv preprint arXiv:2306.00650  (2023)

\bibitem{nado2020TTBN}
Nado, Z., Padhy, S., Sculley, D., D'Amour, A., Lakshminarayanan, B., Snoek, J.: Evaluating prediction-time batch normalization for robustness under covariate shift. arXiv preprint arXiv:2006.10963  (2020)

\bibitem{niu2022EATA}
Niu, S., Wu, J., Zhang, Y., Chen, Y., Zheng, S., Zhao, P., Tan, M.: Efficient test-time model adaptation without forgetting. In: ICML. pp. 16888--16905 (2022)

\bibitem{niu2023SAR}
Niu, S., Wu, J., Zhang, Y., Wen, Z., Chen, Y., Zhao, P., Tan, M.: Towards stable test-time adaptation in dynamic wild world. In: ICLR (2023)

\bibitem{pei2018multi}
Pei, Z., Cao, Z., Long, M., Wang, J.: Multi-adversarial domain adaptation. In: AAAI. vol.~32 (2018)

\bibitem{peng2019moment}
Peng, X., Bai, Q., Xia, X., Huang, Z., Saenko, K., Wang, B.: Moment matching for multi-source domain adaptation. In: ICCV. pp. 1406--1415 (2019)

\bibitem{peng2019domainnet}
Peng, X., Bai, Q., Xia, X., Huang, Z., Saenko, K., Wang, B.: Moment matching for multi-source domain adaptation. In: ICCV. pp. 1406--1415 (2019)

\bibitem{purushotham2016variational}
Purushotham, S., Carvalho, W., Nilanon, T., Liu, Y.: Variational recurrent adversarial deep domain adaptation. In: ICLR (2016)

\bibitem{rusak2022imagenetd}
Rusak, E., Schneider, S., Gehler, P.V., Bringmann, O., Brendel, W., Bethge, M.: Imagenet-d: A new challenging robustness dataset inspired by domain adaptation. In: ICML 2022 Shift Happens Workshop (2022)

\bibitem{sanyal2023domain}
Sanyal, S., Asokan, A.R., Bhambri, S., Kulkarni, A., Kundu, J.N., Babu, R.V.: Domain-specificity inducing transformers for source-free domain adaptation. In: ICCV. pp. 18928--18937 (2023)

\bibitem{schneider2020TTBN}
Schneider, S., Rusak, E., Eck, L., Bringmann, O., Brendel, W., Bethge, M.: Improving robustness against common corruptions by covariate shift adaptation. In: NeurIPS. vol.~33, pp. 11539--11551 (2020)

\bibitem{shu2018dirt}
Shu, R., Bui, H.H., Narui, H., Ermon, S.: A dirt-t approach to unsupervised domain adaptation. In: ICLR (2018)

\bibitem{song2023ecotta}
Song, J., Lee, J., Kweon, I.S., Choi, S.: Ecotta: Memory-efficient continual test-time adaptation via self-distilled regularization. In: CVPR. pp. 11920--11929 (2023)

\bibitem{storkey2009domainshift}
Storkey, A., et~al.: When training and test sets are different: characterizing learning transfer. Dataset shift in machine learning  \textbf{30}(3-28), ~6 (2009)

\bibitem{sun2016deep}
Sun, B., Saenko, K.: Deep coral: Correlation alignment for deep domain adaptation. In: ECCV. pp. 443--450 (2016)

\bibitem{sun2020TTT}
Sun, Y., Wang, X., Liu, Z., Miller, J., Efros, A., Hardt, M.: Test-time training with self-supervision for generalization under distribution shifts. In: ICML. pp. 9229--9248 (2020)

\bibitem{tang2023consistency}
Tang, L., Li, K., He, C., Zhang, Y., Li, X.: Consistency regularization for generalizable source-free domain adaptation. In: ICCV. pp. 4323--4333 (2023)

\bibitem{wang2021tent}
Wang, D., Shelhamer, E., Liu, S., Olshausen, B., Darrell, T.: Tent: Fully test-time adaptation by entropy minimization. In: ICLR (2021)

\bibitem{wang2022exploring}
Wang, F., Han, Z., Gong, Y., Yin, Y.: Exploring domain-invariant parameters for source free domain adaptation. In: CVPR. pp. 7151--7160 (2022)

\bibitem{wang2022CoTTA}
Wang, Q., Fink, O., Van~Gool, L., Dai, D.: Continual test-time domain adaptation. In: CVPR. pp. 7201--7211 (2022)

\bibitem{wilson2020survey}
Wilson, G., Cook, D.J.: A survey of unsupervised deep domain adaptation. ACM Transactions on Intelligent Systems and Technology  \textbf{11}(5),  1--46 (2020)

\bibitem{wu2023metagcd}
Wu, Y., Chi, Z., Wang, Y., Feng, S.: Metagcd: Learning to continually learn in generalized category discovery. In: IEEE International Conference on Computer Vision (2023)

\bibitem{wu2024test}
Wu, Y., Chi, Z., Wang, Y., Plataniotis, K.N., Feng, S.: Test-time domain adaptation by learning domain-aware batch normalization. In: AAAI Conference on Artificial Intelligence (2024)

\bibitem{resxnet}
Xie, S., Girshick, R., Dollar, P., Tu, Z., He, K.: Aggregated residual transformations for deep neural networks. In: CVPR (July 2017)

\bibitem{yang2023conditional}
Yang, E.H., Hamidi, S.M., Ye, L., Tan, R., Yang, B.: Conditional mutual information constrained deep learning for classification. arXiv preprint arXiv:2309.09123  (2023)

\bibitem{Yang2024how}
Yang, E.H., Ye, L.: How to train the teacher model for effective knowledge distillation. In: European Conference on Computer Vision. Springer (2024)

\bibitem{Yang2024Markov}
Yang, E.H., Ye, L.: Markov knowledge distillation: Make nasty teachers trained by self-undermining knowledge distillation fully distillable. In: European Conference on Computer Vision. Springer (2024)

\bibitem{yang2022attracting}
Yang, S., Jui, S., van~de Weijer, J., et~al.: Attracting and dispersing: A simple approach for source-free domain adaptation. In: NeurIPS. vol.~35, pp. 5802--5815 (2022)

\bibitem{yang2021generalized}
Yang, S., Wang, Y., Van De~Weijer, J., Herranz, L., Jui, S.: Generalized source-free domain adaptation. In: ICCV. pp. 8978--8987 (2021)

\bibitem{ye2024bayes}
Ye, L., Hamidi, S.M., Tan, R., YANG, E.H.: Bayes conditional distribution estimation for knowledge distillation based on conditional mutual information. In: The Twelfth International Conference on Learning Representations (2024)

\bibitem{NEURIPS2020conflict}
Yu, T., Kumar, S., Gupta, A., Levine, S., Hausman, K., Finn, C.: Gradient surgery for multi-task learning. In: NeurIPS. vol.~33, pp. 5824--5836 (2020)

\bibitem{yuan2023RoTTA}
Yuan, L., Xie, B., Li, S.: Robust test-time adaptation in dynamic scenarios. In: CVPR. pp. 15922--15932 (2023)

\bibitem{widresnet}
Zagoruyko, S., Komodakis, N.: Wide residual networks. In: Wilson, R.C., Hancock, E.R., Smith, W.A.P. (eds.) BMVC (2016)

\bibitem{2022MEMO}
Zhang, M., Levine, S., Finn, C.: Memo: Test time robustness via adaptation and augmentation. In: NeurIPS. vol.~35, pp. 38629--38642 (2022)

\bibitem{zhang2023long_tailed}
Zhang, Y., Kang, B., Hooi, B., Yan, S., Feng, J.: Deep long-tailed learning: A survey. IEEE TPAMI  \textbf{45}(9),  10795--10816 (2023)

\bibitem{zhao2023delta}
Zhao, B., Chen, C., Xia, S.T.: Delta: Degradation-free fully test-time adaptation. In: ICLR (2023)

\bibitem{zhong2022meta}
Zhong, T., Chi, Z., Gu, L., Wang, Y., Yu, Y., Tang, J.: Meta-dmoe: Adapting to domain shift by meta-distillation from mixture-of-experts. Advances in Neural Information Processing Systems  (2022)

\bibitem{zhou2023ods}
Zhou, Z., Guo, L.Z., Jia, L.H., Zhang, D., Li, Y.F.: Ods: Test-time adaptation in the presence of open-world data shift. In: ICML (2023)

\end{thebibliography}

\end{document}